# Vessel Segmentation and Catheter Detection in X-Ray Angiograms Using Superpixels

Hamid R. Fazlali[1], Nader Karimi[2], S.M. Reza Soroushmehr[3,4], Shahram Shirani[1], Brahmajee.K. Nallamothu[4], Kevin R. Ward[3], Shadrokh Samavi[1,2], Kayvan Najarian[3,5]

[1]Department of Electrical and Computer Engineering, McMaster University, Hamilton, Canada.
[2]Department of Electrical and Computer Engineering, Isfahan University of Technology, Isfahan, 84156-83111, Iran.
[3]Michigan Center for Integrative Research in Critical Care, University of Michigan, Ann Arbor, U.S.A.
[4]Department of Emergency Medicine, University of Michigan, Ann Arbor, U.S.A.
[5]Department of Computational Medicine and Bioinformatics, University of Michigan, Ann Arbor, U.S.A.

**Abstract** Coronary artery disease (CAD) is the leading causes of death around the world. One of the most common imaging methods for diagnosing this disease is X-ray angiography. Diagnosing using these images is usually challenging due to non-uniform illumination, low contrast, presence of other body tissues, presence of catheter etc. These challenges make the diagnoses task of cardiologists tougher and more prone to misdiagnosis. In this paper we propose a new automated framework for coronary arteries segmentation, catheter detection and centerline extraction in x-ray angiography images. Our proposed segmentation method is based on superpixels. In this method at first three different superpixel scales are exploited and a measure for vesselness probability of each superpixel is determined. A majority voting is used for obtaining an initial segmentation map from these three superpixel scales. This initial segmentation is refined by finding the orthogonal line on each ridge pixel of vessel region. In this framework we use our catheter detection and tracking method which detects the catheter by finding its ridge in the first frame and traces in other frames by fitting a second order polynomial on it. Also we use the image ridges for extracting the coronary arteries centerlines. We evaluated our method qualitatively and quantitatively on two different challenging datasets and compared it with one of the previous well-known coronary arteries segmentation methods. Our method could detect the catheter and reduced the false positive rate in addition to achieving better segmentation results. The evaluation results prove that our method performs better in a much shorter time.

## 1. Introduction

Coronary artery disease (CAD) is one of the major causes of death in the world. In this disease the coronary arteries wall get narrowed by a substance called plaque. This narrowness prevents blood to reach the heart properly. This narrowness gradually gets more severe and without treatment, if enough blood does not reach to the heart, heart attack may occur. Cardiologists use different imaging modalities such as x-ray angiography (XRA), computed tomography angiography (CTA), magnetic resonance angiography (MRA), optical coherence tomography (OCT) and intravascular ultrasound (IVUS) for diagnosing this disease. However, XRA is taken as a gold standard for assessment of CAD [1]. In this imaging method, the patient lies down on the bed and a thin hallow tube called catheter is inserted through the patient's arteries from groin, neck or arm. When the catheter reaches the intended artery, the contrast agent is injected and usually 2-4 sequences of heart performance are captured. Cardiologists can find the places of the stenosis and determine their severity by watching these videos. Although these XRA images are being used by cardiologists extensively around the world, they are usually very challenging images and this will make cardiologists' works much harder and in some cases these challenges may cause misdiagnosis. Non-uniform illumination, low contrast, low signal to noise ratios, presence of other body tissues and catheter are some of the most common challenges.

In this paper we propose a new automatic framework for coronary arteries segmentation, centerline extraction and catheter detection. This framework consists of three main stages: preprocessing, arteries segmentation, and catheter detection and tracking. In the preprocessing stage, due to the low contrast nature of the XRA input images, the contrast is enhanced and this enhanced contrast image is also used in the final stage for better representing the segmented image. For better detecting the arteries region, a vesselness detection method is used which is based on Hessian matrix analysis. Also for detecting the image ridges for further processing in the next stages, a proper smoothing method is used which uses the contrast enhanced image and the vesselness map. In the arteries segmentation stage an efficient superpixel algorithm is exploited for coronary arteries segmentation. To the best of our knowledge none of the previous coronary arteries segmentation methods has used superpixels for segmentation. The reason for choosing this method for segmentation is that superpixels are a group of pixels that make the segmentation procedure much easier and faster as they fit on borders of major arteries. In this work because of different scales of arteries, we use three different superpixel scales. Using three different superpixel scales, at first a new measure for vesselness probability of each image superpixel is determined and an initial vessel region is obtained for each superpixel scale. Then the superpixels that are on the ridges and have overlap with the initial vessel region are added to the initial vessel region. A majority voting mechanism is used for each pixel among three vessel regions and an initial segmentation is generated. In order to refine the initial segmentation result, orthogonal lines on each arteries ridge pixels are found and by detecting the arteries borders, other parts are removed. Also, in this framework in order to find the catheter and reduce our segmentation false positive errors, in images that the catheter exists, we use our previous catheter detection and tracking method [2]. This method is based on image ridges and by finding the catheter ridge in the first frame of a sequence and fitting a second order polynomial on it, it is tracked through the sequence. For extracting the arteries centerlines, we calculate the image ridges and by multiplying this image ridge with the segmentation mask, the centerlines are produced.

The organization of the remainder of this paper is as follows. In Section 2, some of the related works are reviewed. In Section 3, background concepts are explained. The proposed method is presented in Section 4 and the quantitative and qualitative results of the proposed arteries segmentation and catheter detection and tracking are presented in Section 5. Discussions on the obtained results are presented in Section 6. Finally, in Section 7 the paper is concluded.

## 2. Related Works

During the recent years many methods have been proposed for XRA images. These methods can be categorized into two parts as coronary arteries enhancement methods and coronary arteries segmentation methods. In the following we will review some of the enhancement and segmentation methods. The aim of the enhancement methods is to improve visual quality of the input XRA image which may help easier segmentation of arteries. In [3], by constructing the Hessian matrix of the input image and analyzing its eigenvalues, a vesselness measure is obtained. This filter which is known as Hessian filter is widely used in enhancement and segmentation of coronary arteries. Because of some drawbacks of this method such as noise sensitivity and junction suppression, in [4], a method for vesselness measurement is proposed which is based on decimation-free directional filter bank. In this method, at first the input image is decomposed into directional images. After applying Homomorphic filter on each directional image, the vesselness measure is obtained for each directional image. Finally all the directional images are averaged and the final vesselness image is obtained. In [5], at first the varying illumination is removed and after normalizing the illuminated image, it is decomposed into several directional images. Finally a high pass filter is applied on each directional images and each block of the output image is constructed by selecting a coordinated block from those directional images with the highest energy.

In [6], a method for detecting and enhancing the ROI in XRA images is proposed. In this method the vesselness measure of Hessian filter is enhanced using Guided filter. Then multi-scale multi-threshold Canny edge detector is used for detecting the arteries edges in the enhanced vesselness map. In the last stage, the ROI region is detected by searching each block of the edge image in an overlap format and the background area is smoothed. Since the diaphragm border is present in some XRA images and may mislead the artery segmentation method, in [7], a diaphragm border detection and removal method is proposed. In this method at first all vessels are removed by using closing morphological operator. Then the edges that their upsides are brighter than their downside are detected. By using an energy function, a polynomial is fitted on the edges of the diaphragm. In [8] the authors proposed a method that is able to separate an X-ray angiogram into three separate layers which are breathing layer, a quasi-static layer and a vessel layer. These layers were extracted by using morphological closing operator and an online PCA algorithm.

Another method for increasing XRA image quality is through using digital subtraction angiography (DSA). In this method a frame at the beginning of an XRA sequence, in which the contrast agent is not injected, is considered as a mask frame. Also, a frame, in which the contrast agent is fully injected, is considered as the contrast frame. Then, the mask frame is subtracted from the contrast frame. In an ideal sequence, in which there is no camera and heart movements, in the resulted image the background is omitted and the arteries are enhanced. But due to the camera and heart motions, some artifacts will be generated due to this simple mask subtraction technique.

In order to deal with DSA artifacts, several methods have been proposed for registration of DSA images. In [9], a three steps method is proposed. In this method at first a contrast image is selected and an image most similar to the selected contrast image is selected using energy of the histogram of the differences (EHD). Then the B-spline based free-form deformation non-rigid image registration is used for registration of the contrast and mask images. In the last step, independent component analysis (ICA) is used for separation of the arteries and the background. In [10], a multi-resolution algorithm for non-rigid image registration is proposed, in which the input images are decomposed into coarse and fine sub-image blocks iteratively. All the corresponding block pairs at each decomposition level are affinely registered in a multi-scale framework. Explicit changes of the contrast and brightness are also incorporated into the registration model. These local transformations are then smoothly interpolated using thin-plate spline to obtain the global model. Due to severe camera or heart motions these registration methods cannot enhance the image quality.

Some researches in recent years have been devoted to coronary arteries segmentation methods. In [11], several vessel segmentation methods are reviewed and categorized as pattern recognition, model-based, tracking-based, artificial intelligence-based, neural network-based, and miscellaneous tube-like object detection approaches. In [1], another review is done on coronary arteries segmentation methods and their capabilities are compared. In [12] the vesselness measure obtained by Hessian filter was combined with flux flow measurements and the vessels were segmented by analyzing the connected components. In [13], an active contour-based segmentation method is proposed. In this method the initial contour is determined using Hessian filter and according to the degree of homogeneity in the image, the local and global forces are automatically weighted. In [14], using directional Hessian-based method [4], the vesselness measure is computed and is used in the local feature fitting energy function for forcing the contour to segment the arteries.

In [15], a graph-cut based segmentation method is proposed for coronary arteries. In this method the initial background and arteries pixels are selected automatically using ridge detection and Hessian filter respectively. Then because of low probability of some vessel parts in Hessian filter, geodesic vesselness measure is used. This modified vesselness measure is used in the unary term in the graph-cut energy function. Multi-scale multi-threshold Canny is used in the boundary term of the vesselness function. In [17], after detecting several frames in a sequence as high-contrast angiograms, the Hessian filter result is enhanced. Then by binarizing the obtained vesselness map, the largest connected component is selected as arteries component and connected-component analysis is used for detecting the connected regions in the vessel map.

## 3. Background

In this section the background concepts of our method are discussed. These concepts include superpixel, vesselness map and contrast enhancement and smoothing that are exploited in our proposed framework.

### *A. Superpixel*

Superpixel algorithms are segmentation methods used for grouping pixels into perceptually meaningful atomic regions which can be used to replace the rigid structure of the pixel grid [16]. Simple linear iterative clustering (SLIC) [16] method is a fast and efficient superpixel algorithm. In this method at first a desired superpixel number $k$ is determined on the image with $N$ pixels. Then a superpixel grid is produced with $S = \sqrt{N/K}$ intervals. These superpixel centers are moved in a $3 \times 3$ window in order to place them in the lowest gradient position. Then each pixel $i$ in the image is assigned to the nearest cluster center whose search region overlaps it. This is done by computing a distance measure shown in (1):

$$D = \sqrt{d_c^2 + (\frac{d_s}{S})^2\, m^2} \tag{1}$$

In (1), $D$ is the distance between a pixel $i$ and a cluster center $c_j$, $m$ shows the relative importance between color similarity and spatial proximity, $d_c$ and $d_s$ are color distance and spatial distance which are shown in (2) and (3) respectively:

$$d_c = \sqrt{(l_j - l_i)^2} \tag{2}$$

$$d_s = \sqrt{(x_j - x_i)^2 + (y_j - y_i)^2} \tag{3}$$

In (2) and (3), $l_i, x_i$ and $y_i$ represent the intensity value, vertical and horizontal positions of pixel $i$ respectively. After assigning each pixel to a cluster center, an update step changes the cluster centers to the mean $[l, x, y]^T$ vector of all pixels belonging to the cluster. This assignment and updating steps are repeated iteratively until the error, which is computed using $L_2$ norm, converges.

By using SLIC superpixel method on XRA images, because of the intensity differences between artery parts and background parts, the superpixels are accurately fit on the arteries borders. This appropriate fitness of superpixels on artery borders, mainly the major arteries, makes the segmentation procedure much easier. The only and the important challenge that is remaining is distinguishing the arteries superpixels from the background superpixels. This will be addressed by our proposed segmentation method.

### *B. Vesselness Map*

Vesselness-measure is calculated for every pixel showing the probability of that pixel being a vessel pixel. Vesselness measure could be obtained by using Hessian filter. In this filter at first the Hessian matrix of the input image is constructed as:

$$H = \begin{bmatrix} I_{xx} & I_{xy} \\ I_{xy} & I_{yy} \end{bmatrix} \tag{4}$$

where, $I_x$ and $I_y$ are computed by convolving the first order derivatives of Gaussian function in horizontal and vertical directions as shown in (5) and (6):

$$I_x = \sigma^l G_{x,\sigma} * I \tag{5}$$

$$I_y = \sigma^l G_{y,\sigma} * I \tag{6}$$

In (5) and (6), $l$ is the Lindberg factor used for normalizing the images derivative in different scales, which is 1 in this method. Also, $*$ is convolution sign, $G_{x,\sigma}$ and $G_{y,\sigma}$ are first order derivatives in $x$ and $y$ directions of the Gaussian function with standard deviation of $\sigma$ respectively. By considering two eigenvalues of the Hessian matrix as $\lambda_1$ and $\lambda_2$ that $\lambda_1 < \lambda_2$, the vesselness measure for scale $s$ is computed using (7):

$$V_o(s) = \begin{cases} 0 & if \lambda_2 < 0 \\ e^{-\frac{R_B{}^2}{2\beta^2}} \left(1 - e^{-\frac{S^2}{2c^2}}\right) & if \lambda_2 \geq 0 \end{cases} \tag{7}$$

where and $S = \sqrt{\lambda_1^2 + \lambda_2^2}$ is the structural similarity measure and $R_B = \lambda_1/\lambda_2$ is the non-similarity measure. Also $\beta$ and $c$ are two thresholds that control the non-similarity and the structural similarity measures respectively. The Hessian final vesselness measure is obtained by using all scales as follows:

$$V = \max_{s_{min} \leq s \leq s_{max}} V_o(s) \tag{8}$$

In (8), $V$ is the final vesselness measure and $s_{min}$ and $s_{max}$ are minimum and maximum scales used for vessels.

Because of noise sensitivity and junction suppression of Hessian filter, in [4], decimation-free directional filter bank (DDFB) is used. In [4] at first the input image is decomposed into 8 directional images. This decomposition helps avoiding junction suppression and the noise in each directional image. Due to the non-uniform illumination of these images, homomorphic filter is applied on the corresponding direction of each directional image. Then Hessian filter is applied on each directional image. In order to prevent image degradation due to alignment of vessels with x-axis, the coordinates are rotated. Hence, the Hessian matrix is changed as follows:

$$H' = \begin{bmatrix} I_{x'x'} & I_{x'y'} \\ I_{x'y'} & I_{y'y'} \end{bmatrix} \quad (9)$$

where

$$I_{x'x'} = I_{xx}cos^2\theta_i + I_{xy}\sin(2\theta_i) + I_{yy}sin^2\theta_i \quad (10)$$

$$I_{y'y'} = I_{xx}sin^2\theta_i + I_{xy}\sin(2\theta_i) + I_{yy}cos^2\theta_i \quad (11)$$

$$I_{x'y'} = -\frac{1}{2}I_{xx}\sin(2\theta_i) + I_{xy}\cos(2\theta_i) + \frac{1}{2}I_{yy}\sin(2\theta_i) \quad (12)$$

In (10), (11) and (12), $\theta_i$ is the orientation of a directional image with minimum and maximum orientations of $\theta_{i,min}$ and $\theta_{i,max}$ which is computed as:

$$\theta_i = \frac{\theta_{i,min} + \theta_{i,max}}{2} \quad (13)$$

*C. Contrast Enhancement*

Top-hat transform is a morphological based contrast enhancement method, which increases the difference between the bright and dark regions of an image. Due to the different scales of arteries in XRA images, multi-scale top-hat transform can increase their contrast properly. In [18], top-hat transform was used in a multi-scale format. This multi-scale transform consists of two essential morphological operators, which are opening and closing, for extracting the bright and dark regions in different scales as follows:

$$BTH_i = I - IoB_i \quad (14)$$

$$DTH_i = I \bullet B_i - I \quad (15)$$

In (14) and (15), $BTH$ and $DTH$ are bright and dark regions extracted in scale $i$ and $I$ is the input image. Also, $B$ is a structuring element with the size of $i$ and o and • are opening and closing morphological operators respectively. These dark and bright extracted regions are used for producing the final dark and bright regions. For this aim, the maximum bright regions are computed using different bright scales. The result is then added to the maximum of the differences of consecutive bright regions. The same process is done for dark regions. Finally the contrast-enhanced image is produced by adding the multi-scale bright region to the original image. Then the multi-scale dark region is subtracted from the original image as shown below:

$$I_{CE} = I + I_w - I_B \quad (16)$$

In (16), $I_{CE}$ is the contrast enhanced image using multi-scale top-hat transform, $I$ is the original image, $I_w$ and $I_B$ are multi-scale dark and bright regions.

*D.Smoothing*

Due to the presence of noise in images, different smoothing methods are used. One of the most prominent methods for smoothing is the use of Guided filter [19]. Guided filter is an edge-preserving smoothing filter. Guided filter uses two images known as guided image $F$ and filtering input image $p$. This filter assumes that the smoothed output image $q$ is a linear model of the guidance image in a window $\omega_k$ with radius $r$ centered in pixel $k$ as follows:

$$q_i = \alpha_k F_i + \beta_k, \qquad \forall i \in \omega_k \quad (17)$$

In order to find the constant coefficients ($\alpha_k$,$\beta_k$) in $\omega_k$, the guidance and filtering input images are used as shown in (18) and (19):

$$\alpha_k = \frac{\frac{1}{|\omega|}\sum_{i\in\omega_k} F_i p_i - \mu_k \bar{p}_k}{\sigma_k^2 + \epsilon} \quad (18)$$

$$\beta_k = \bar{p}_k - \alpha_k \mu_k \quad (19)$$

In (18) and (19), $\mu_k$ and $\sigma_k$ are average and variance of the guidance image $F$ in $\omega_k$ window and also $\epsilon$ is aregularization parameter which penalizes large $\alpha_k$.

## 4. Proposed Method

In this section, the proposed framework is explained with details. The block diagram of the proposed framework is shown in Fig. 1. As can be seen, our method consists of three major stages. These stages are preprocessing, catheter detection and coronary arteries segmentation. The results of the preprocessing stage are passed to the catheter detection and coronary arteries segmentation stages. After segmenting the coronary arteries, and detecting the catheter and extracting the centerlines, three masks are generated for arteries and catheter. Then they are overlaid on the contrast enhanced image. The details of each stage are discussed in the following.

*A. Preprocessing*

The aim of this stage is the preprocessing of the input image in order to produce a proper image and a corresponding map for segmentation and catheter detection stages.

*1) Contrast Enhancement*

One of the challenging problems in processing XRA images is their low contrast intensity that makes it hard to differentiate vessels from the background. In order to increase the contrast of these images, by considering the scales of arteries, we use multi-scale top-hat transform [18]. In Fig. 2, the result of applying this transform on an XRA image is shown. By comparing Fig. 2(a) and Fig. 2(b), we can see that the contrast of the Fig. 2(a) has properly been increased and the arteries structure is completely visible after contrast enhancement.

*2) Vesselness Measurement*

In order to obtain an appropriate vesselness map, we use the method of [5], which has low noise sensitivity and high robustness in junction areas as compared to Hessian filter. We use this vesselness measure in two distinct steps. The first one is for obtaining a vesselness map from the original input image in order to exploit it for smoothing the input image. The second step is for obtaining a vesselness map from the contrast enhanced image. This is used in the segmentation stage for determining the vesselness of each superpixel. In Fig. 3, the result of applying this vesselness measurement method on an XRA image is shown. By comparing the original image with its vesselness map, we can see that the arteries have higher vesselness values as compared to the background.

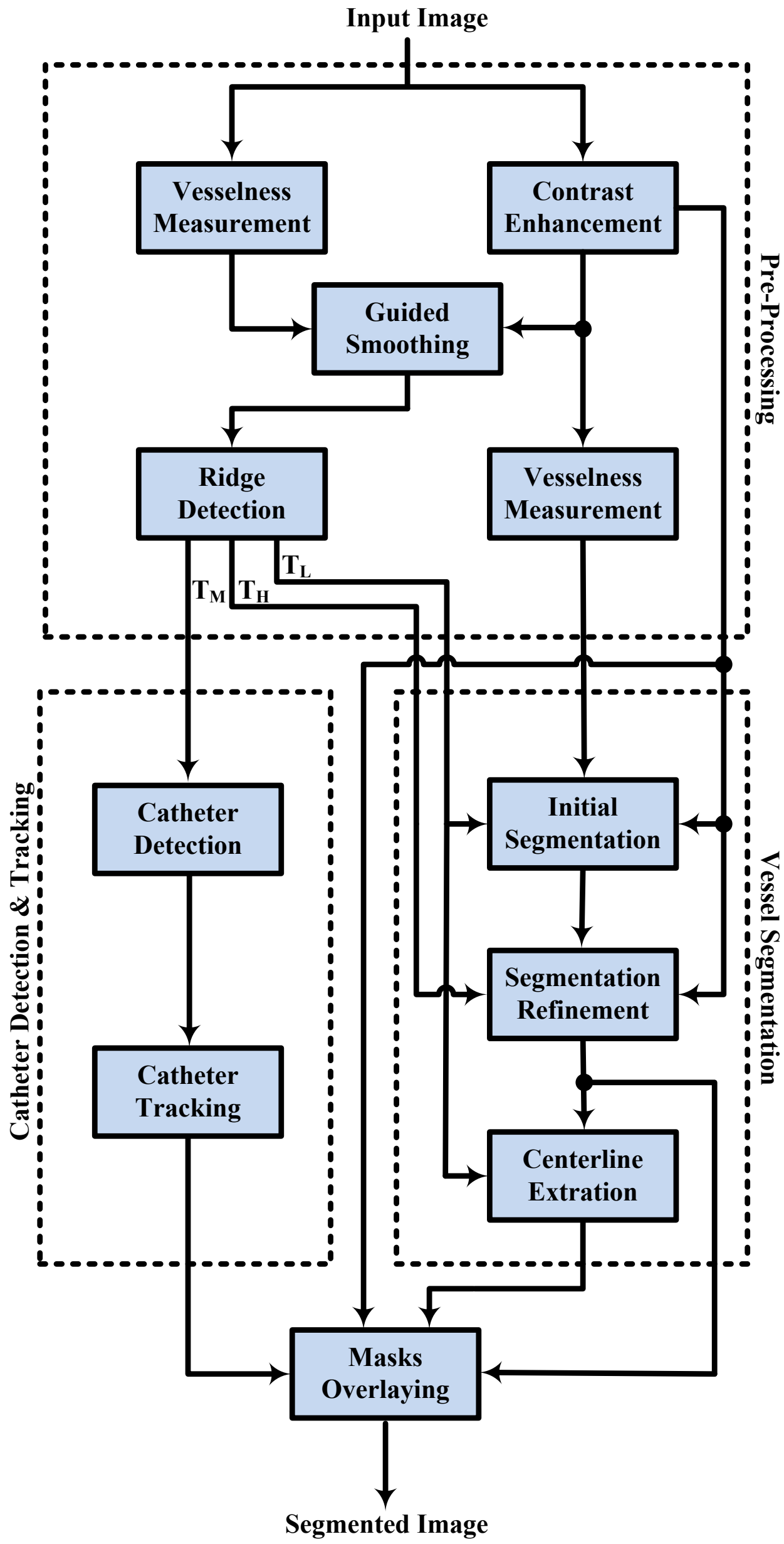

Fig. 1. Block diagram of the proposed method.

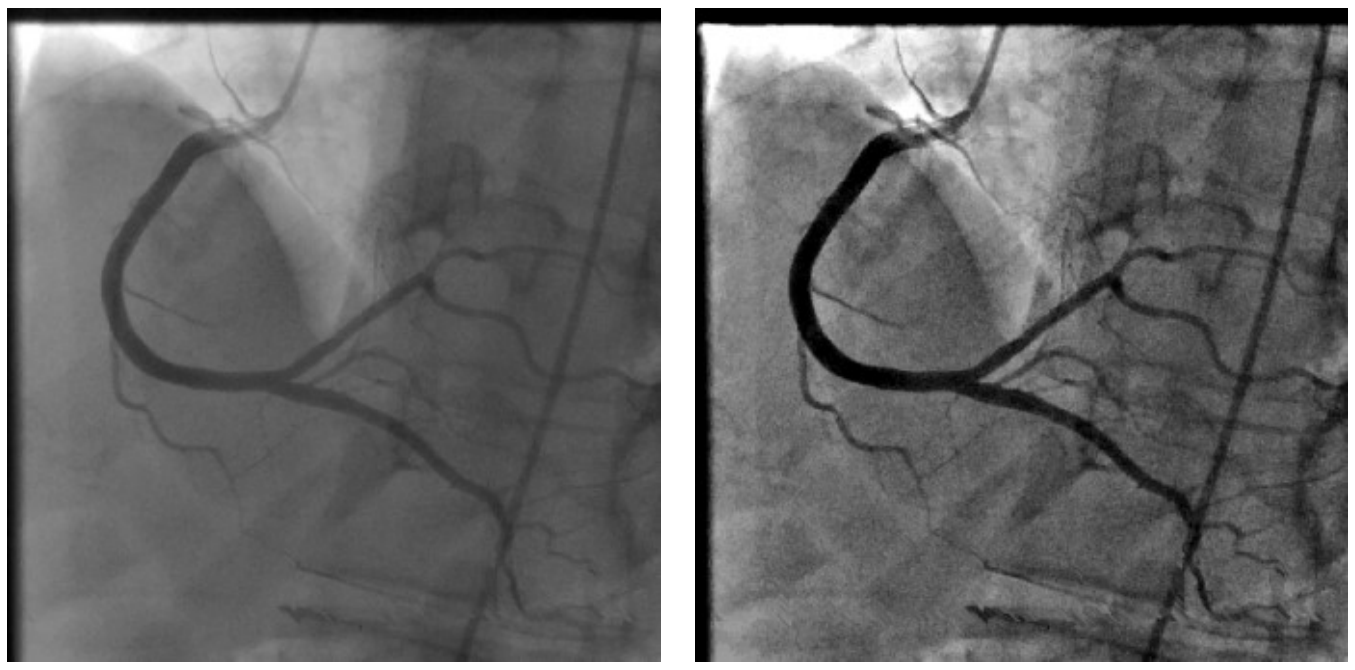

(a) (b)

Fig. 2. Result of contrast enhancement using multi-scale top-hat transform. (a) Original image, (b) contrast enhanced image.

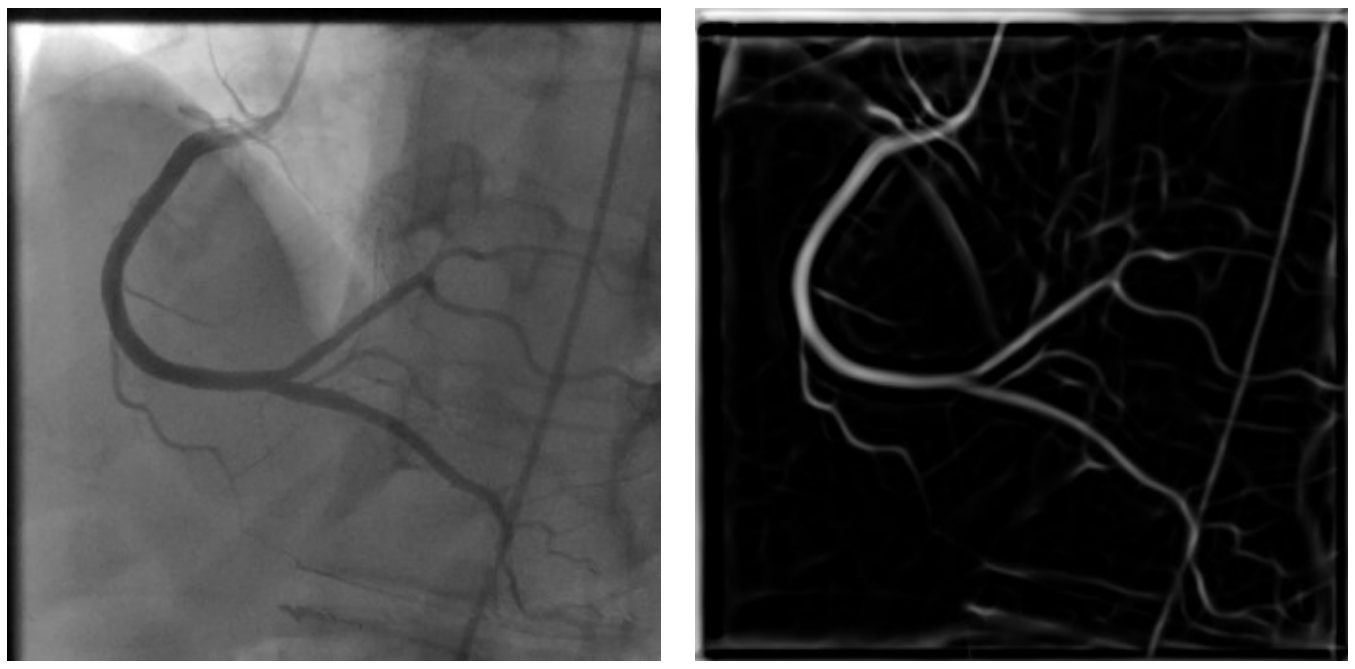

(a) (b)

Fig. 3. Result of vesselness measurement using directional Hessian-based method. (a) Original XRA image, (b) vesselness map.

### *3) Guided Smoothing*

In addition to presence of arteries and catheter structures in the original image and in the contrast enhanced image, other body organs and tissues might exist in these images. As we want to extract image ridges in the next step, in this step we smooth the input image. This is done to remove other body tissues and organs that may add some unwanted and extra ridges. Simple smoothing functions such as Gaussian filter can smooth an image properly. This may also suppress important image edges. Therefore, we use guided filter [19] which is an edge-preserving smoothing filter. In this paper we used the vesselness map as the guidance image and contrasted enhanced image as the filtering input image. Hence, we can smooth the contrast-enhanced image by preserving its important valleys. In Fig. 4, the result of applying this edge-preserving smoothing filter on an XRA image is shown. It can be seen that the arteries and catheter valleys are preserved appropriately and other regions are smoothed.

### *4) Ridge Detection*

Image ridges have important role in our proposed segmentation and catheter detection. As arteries and catheters are exposed as valleys, finding these valleys and their ridges are vital for their detection. In this method we have exploited a ridge detection method based on [20]. In this ridge detection method using Equation (20), valleys that are more than a threshold are obtained and their ridges are determined.

$$Valley = \{j | S_j > \mu_v\} \tag{20}$$

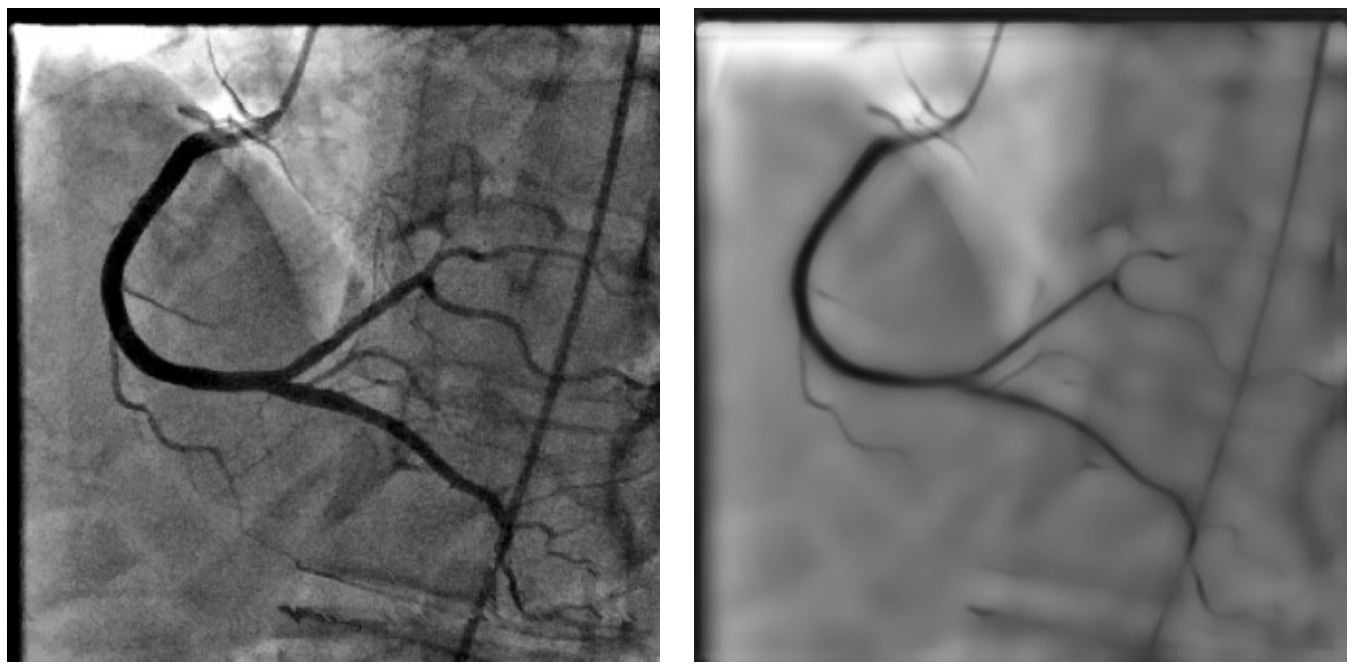

(a) (b)

Fig. 4. Result of applying guided filter. (a) XRA contrast enhanced image, (b) smoothed image.

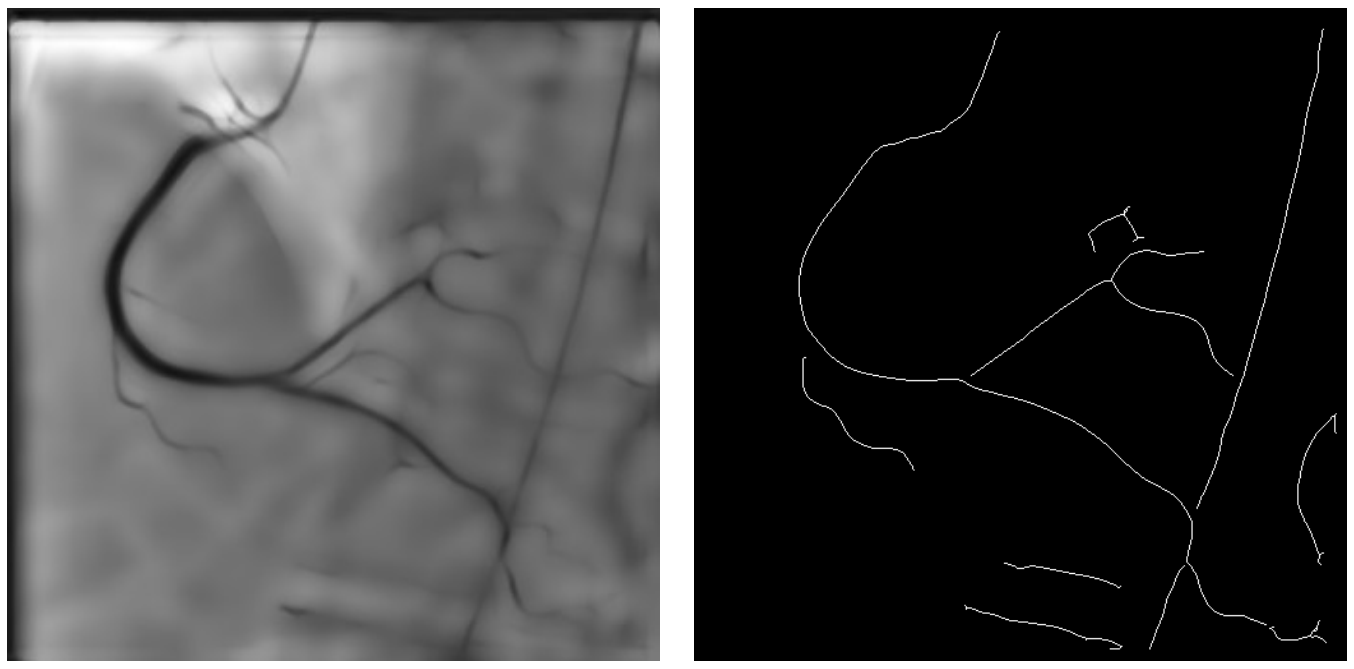

(a) (b)

Fig. 5. Result ridge detection. (a) Smoothed XRA image. (b) Ridge map.

In (20), $Valley$ represents the set of all detected valleys in the smoothed image, $S_j$ is the valley response at pixel $j$ and $\mu_v$ is the threshold for valley detection. Instead of using one threshold, we use three different thresholds. The higher threshold $T_H$, which generates less ridge lines, is used in the segmentation refinement stage. The medium threshold $T_M$, which results in more ridge lines, is used in the catheter detection and tracking stage. The lower threshold $T_L$, which produces more ridges than medium threshold, is used in the initial segmentation and in the centerline extraction stage. In Fig. 5 the result of applying this ridge detection method, on the smoothed image with threshold of 0.2, is shown. It can be seen that the image ridges including artery ridges and catheter ridge are properly detected.

### *B. Catheter Detection and Tracking*

The purpose of this stage is to detect the catheter ridge in the first frame of a sequence. In this frame it is easier to detect and we can track the catheter ridge throughout the sequence. After detection, a mask for catheter is produced that is used for distinguishing the catheter from the arteries.

#### *1) Catheter Ridge Detection*

In the first few frames of an XRA sequence, only catheter and some artifacts exist. But when the contrast agent is injected, artery ridges appear too. Therefore, we detect the catheter ridge in the first frame and then track it throughout the sequence.

In order to detect the catheter ridge, we proposed an algorithm based on Hough transform [2]. In this algorithm, we apply Hough transform on the first ridge frame of the XRA sequence and find the top ten longest line segments that are fit on the ridges. In the ridge map, the catheter ridge appears as the longest curve and more lines are fit on it than other curves produced by background artifacts. Therefore, the ridge segment with the maximum lines on it is selected as the catheter ridge. Also if two ridges exist with the same number of line segments on them, we select the ridge with the longest line. In Fig. 6 the result of this algorithm is shown. Fig. 6(a) is first XRA frame, Fig. 6(b) is the first ridge frame. Figure 6(c) shows the lines fitted on the first frame ridges. Figure 6(d) is the catheter ridge that is selected by our proposed algorithm. By comparing Fig. 6(b) and Fig. 6(d), our proposed algorithm could successfully find the catheter ridge in the first frame.

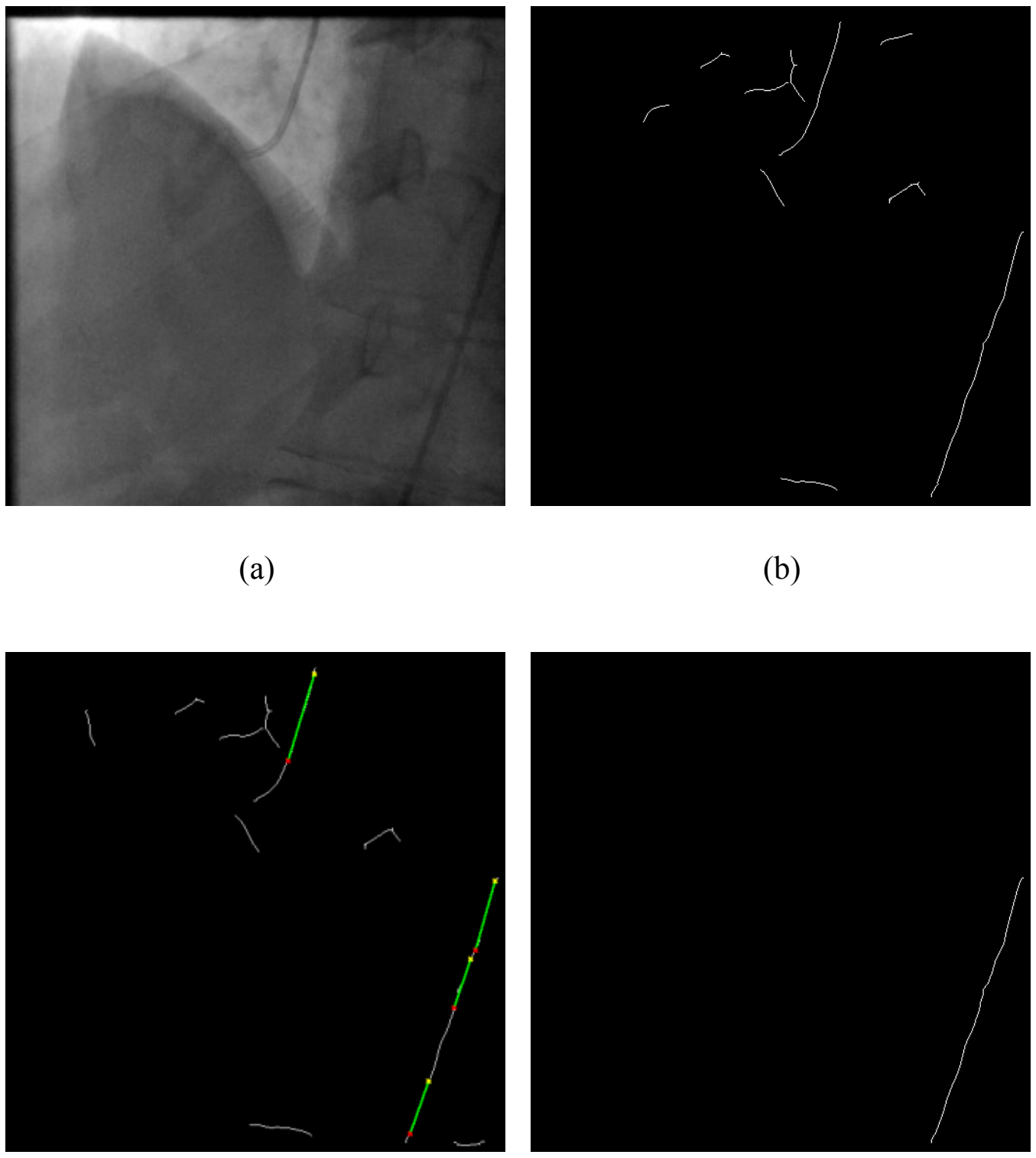

(a) (b) (c) (d)

Fig. 6. Result of Proposed catheter detection algorithm. (a) Original XRA first frame. (b) First frame ridge map. (c) Result of finding lines using Hough transform. (d) Detected catheter ridge.

#### *2) Catheter Tracking*

Due to the camera and heart motions the catheter ridge appears to have a movement in different frames. Therefore, the position of catheter ridge in the second frame differs from its position in the first frame. Hence, we fit a second order polynomial on the catheter ridge in the first frame and use its parameters in order to find the catheter ridge in the second frame. A second order polynomial is as follows:

$$y = ax^2 + bx + c \qquad (21)$$

By fitting this polynomial on the detected catheter ridge in the first frame, we obtain 3 parameters of $a$, $b$ and $c$. We pass these parameters to the second frame and use them as initial values for searching the catheter ridge in limited search area. The second order polynomial, having the most ridge pixels, is detected using Equation (22). This process is done for all frames sequentially.

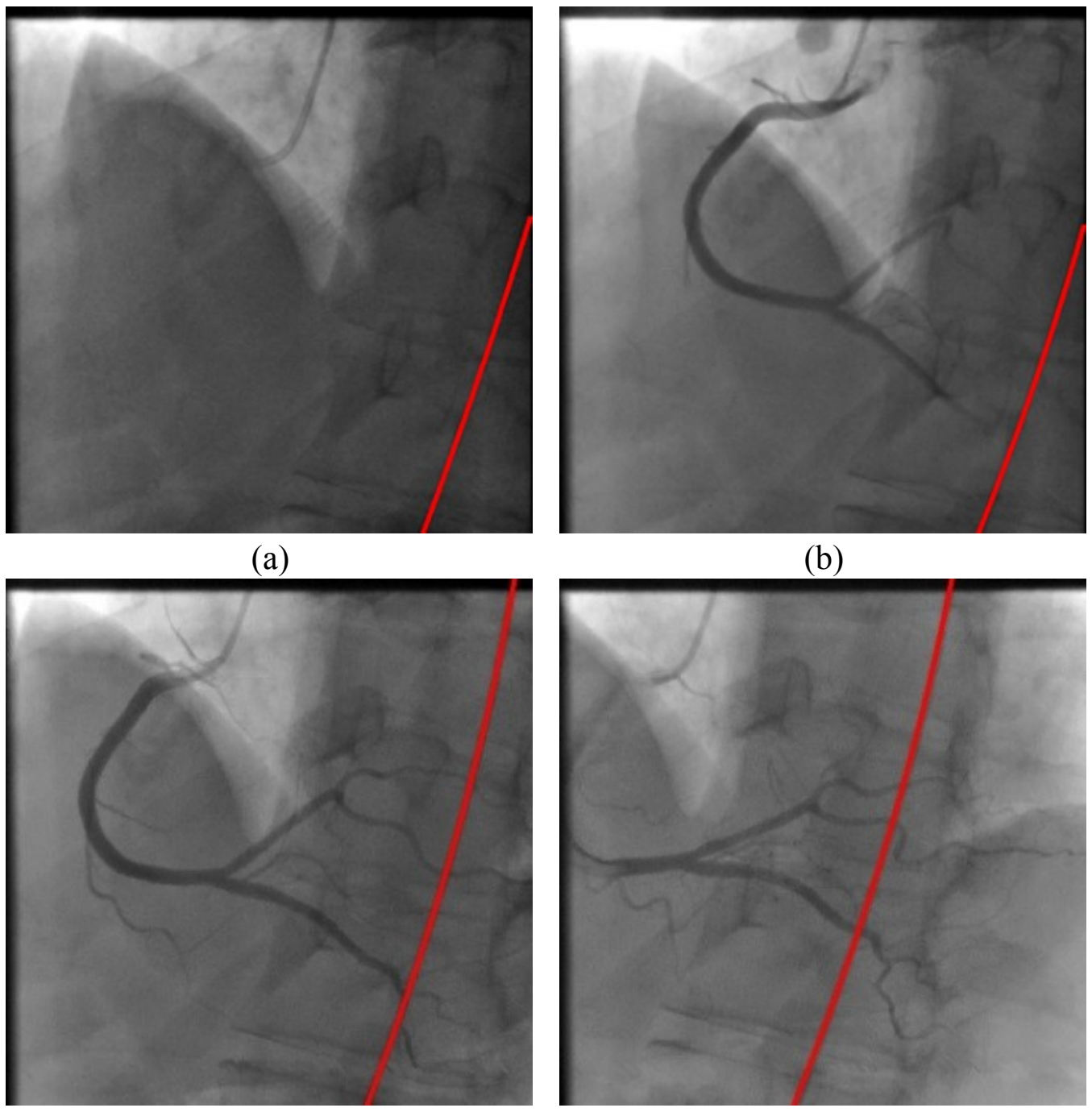

(a) (b) (c) (d)

Fig. 7. Result of proposed catheter tracking algorithm in different phases. (a) Before injection. (b) Beginning of injection. (c) Fully injection. (d) End of injection.

$$BestCurve_i = \underset{a_i, b_i, c_i}{\operatorname{argmax}}\{N(a_i, b_i, c_i)\} \quad (22)$$

In (22) $BestCurve_i$ is the set of coefficients $a$, $b$ and $c$ that best fit on the catheter ridge in frame $i$. Also, $N(a_i, b_i, c_i)$ is the number of ridge pixels that lie on the specified polynomial. In Fig. 7 the results of our catheter detection method on four frames of an XRA sequence are shown. In Fig. 7 four different phases including before injection, beginning of injection, fully injection, end of injection are shown. As can be seen, the proposed method has successfully tracked the catheter and even the presence of arteries and camera and heart motions have not distracted it.

### *C. Coronary Arteries Segmentation*

In this stage we propose our new coronary arteries segmentation method. In this method at first a naïve segmentation is done using superpixel and ridge map. Then the result of this segmentation is refined by finding orthogonal lines on the arteries and removing extra parts including background.

#### *1) Initial Segmentation*

Superpixel algorithms group pixels into several superpixels that are perceptually meaningful. In this paper we exploit SLIC superpixel algorithm [16]. In Fig. 8, the result of applying SLIC superpixel algorithm on a contrast enhanced XRA image is shown. By applying this superpixel algorithm on XRA images, superpixels are completely fit on arteries specially the main arteries.

The only challenge for segmenting the arteries using superpixels is how to find and distinguish the arteries superpixels from those of the background.

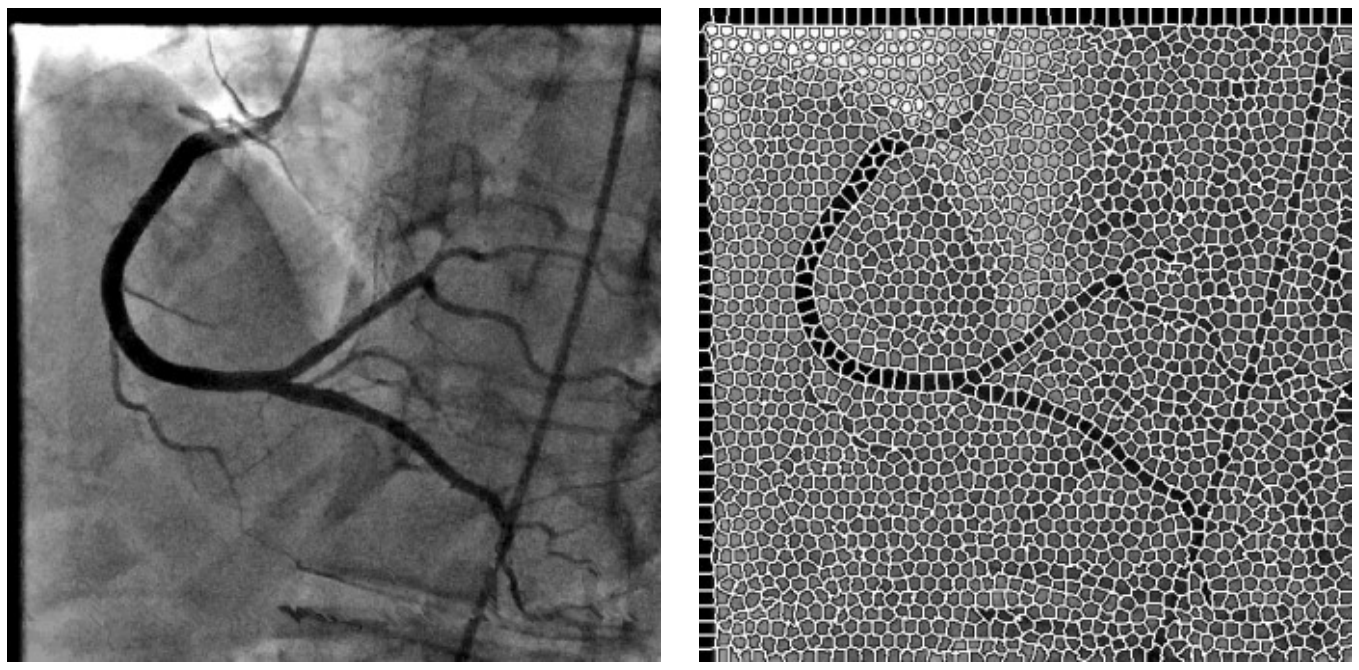

(a) (b)

Fig. 8. Result of applying SLIC superpixel algorithm on a contrast enhanced XRA image. (a) Contrast enhanced XRA image, (b) SLIC result.

For doing so, we use both the contrast enhanced image and the vesselness map, which is obtained from the preprocessing stage. At first the SLIC superpixel algorithm is applied on the contrast enhanced image, as illustrated in Fig. 8. Then the computed superpixel grid is laid on the corresponding vesselness map. By doing this, for each superpixel in the contrast enhanced image, we have a corresponding superpixel in the vesselness map. Let $I_{ce}$ and $I_v$ be the contrast enhanced image and the vesselness map respectively. Then the average intensities of every two corresponding superpixels in $I_{ce}$ and $I_v$ are computed as:

$$n_{ce,s_i} = \frac{\{\sum_j s_i(j) \,|s_i \in I_{ce}\}}{N_{s_i}} \quad (23)$$

$$n_{v,s_i} = \frac{\{\sum_j s_i(j) \,|s_i \in I_{v}\}}{N_{s_i}} \quad (24)$$

where $n_{ce,s_i}$ and $n_{v,s_i}$ are the average intensities of the $i$th superpixels in the contrast enhanced image and vesselness map respectively. Also, $j$ and $N_{s_i}$ are the $j$th pixel and the number of pixels in the $i$th superpixel respectively.

As shown in Fig. 9, artery pixels are darker than the background pixels in the contrast enhanced image. This situation is opposite in the vesselness map where artery pixels are brighter than the background pixels. Therefore the value of $n_{en,s_i}$ should be low in artery superpixels and it should be high in background superpixels. On other hand, the value of $n_{v,s_i}$ should be high in artery superpixels and low in background superpixels.

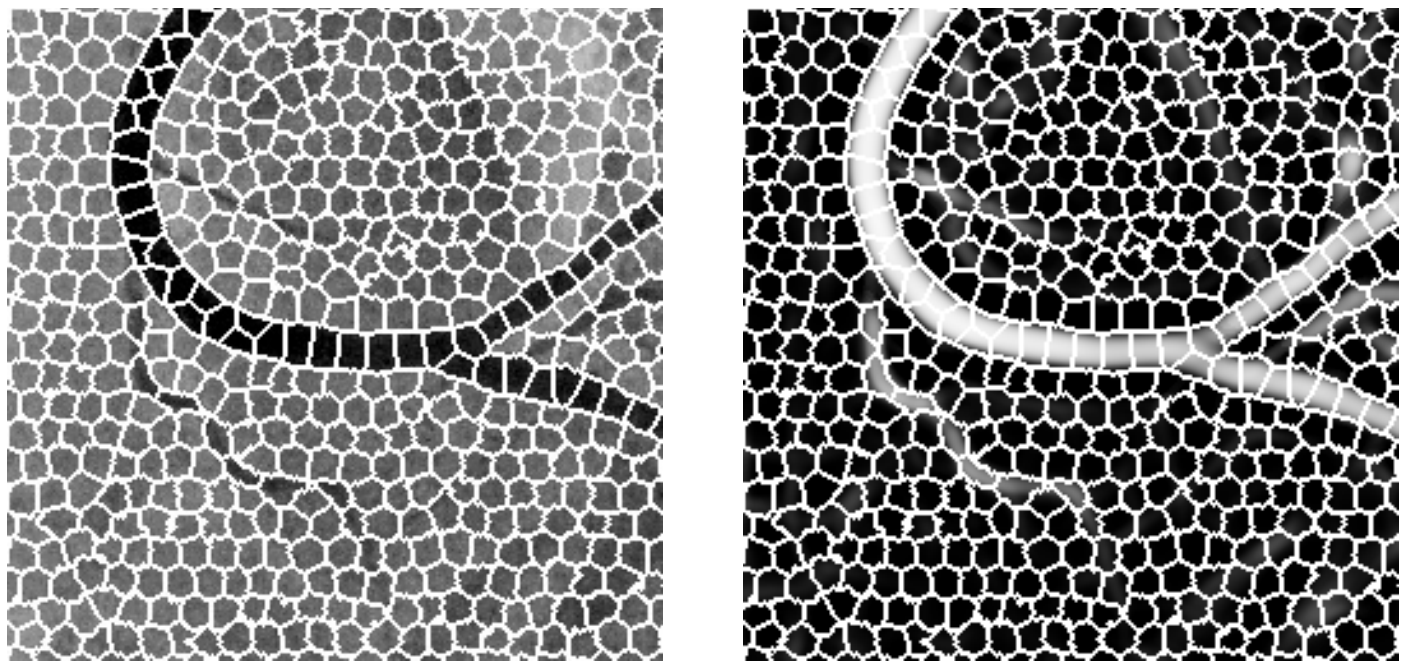

(a) (b)

Fig. 9. Comparing superpixels intensity in contrast enhanced image and its vesselness map. (a) Contrast enhanced XRA image, (b) obtained vesselness map.

In order to distinguish artery superpixel from background superpixels we compute a vesselness probability for each superpixel as follows:

$$\rho_{s_i} = n_{v,s_i} - n_{ce,s_i} \tag{25}$$

where $\rho_{s_i}$ is the vesselness probability for the $i$th superpixel. The value of $\rho_{s_i}$ in (25) is normalized between 0 and 1. After normalization in the artery superpixels, due to the large values of $n_{v,s_i}$ and small values of $n_{en,s_i}$, the $\rho_{s_i}$ values should be high (near 1). In the background superpixels, due to the small values of $n_{v,s_i}$ and large values of $n_{en,s_i}$, the $\rho_{s_i}$ values should be low (near 0). By doing so, a vesselness probability is obtained for each superpixel. In order to find artery superpixels we determine a threshold for $\rho_{s_i}$ values as illustrated in (26):

$$\begin{cases} s_i \in Vessel & if\ \rho_{s_i} \geq T \\ s_i \in Background & if\ \rho_{s_i} < T \end{cases} \tag{26}$$

where $s_i$ is the $i$th superpixel and $T$ is the threshold. Therefore the superpixels, with $\rho_{s_i}$ values higher than $T$, are determined as arteries. Also, superpixels with $\rho_{s_i}$ lower than $T$ are considered as background. It worth mentioning that a very low $T$ value will cause a large part of the background to be selected as arteries and a high $T$ value will dismiss major artery parts. Therefore, a tradeoff should be considered.

In order to make the initial segmentation better, we use the low threshold ($T_L$) ridge map obtained in the preprocessing stage. We find ridges that have overlap with the segmented regions using superpixels. Any superpixel, which lies on these overlapping ridges, is added to the segmentation result.

In XRA images arteries with different scales are present. Hence, working with just one scale of superpixels is problematic and we may lose major parts of arteries. Hence, the whole procedure is performed for three different superpixel scales. Then three initial segmentation maps are obtained from these scales and they are used for generating a unique initial segmentation result. In order to do this, we exploit majority voting between each corresponding pixel in these three maps. In this way, for each pixel, if more than one map votes to its vesselness, the pixel is considered as vessel and otherwise it will be treated as background. In Fig. 10, the last initial segmentation results on two different XRA images are shown.

By comparing the segmentation results with the original XRA images, it can be seen that the initial results have completely segmented the arteries. In the initial segmentation results, in addition to the arteries, some background superpixels are also selected as arteries due to having $\rho_{s_i}$ values (as some background parts resemble the arteries). Because of this, in order to remove these extra parts, we need to do some refinements on the initial segmentation obtained in this step.

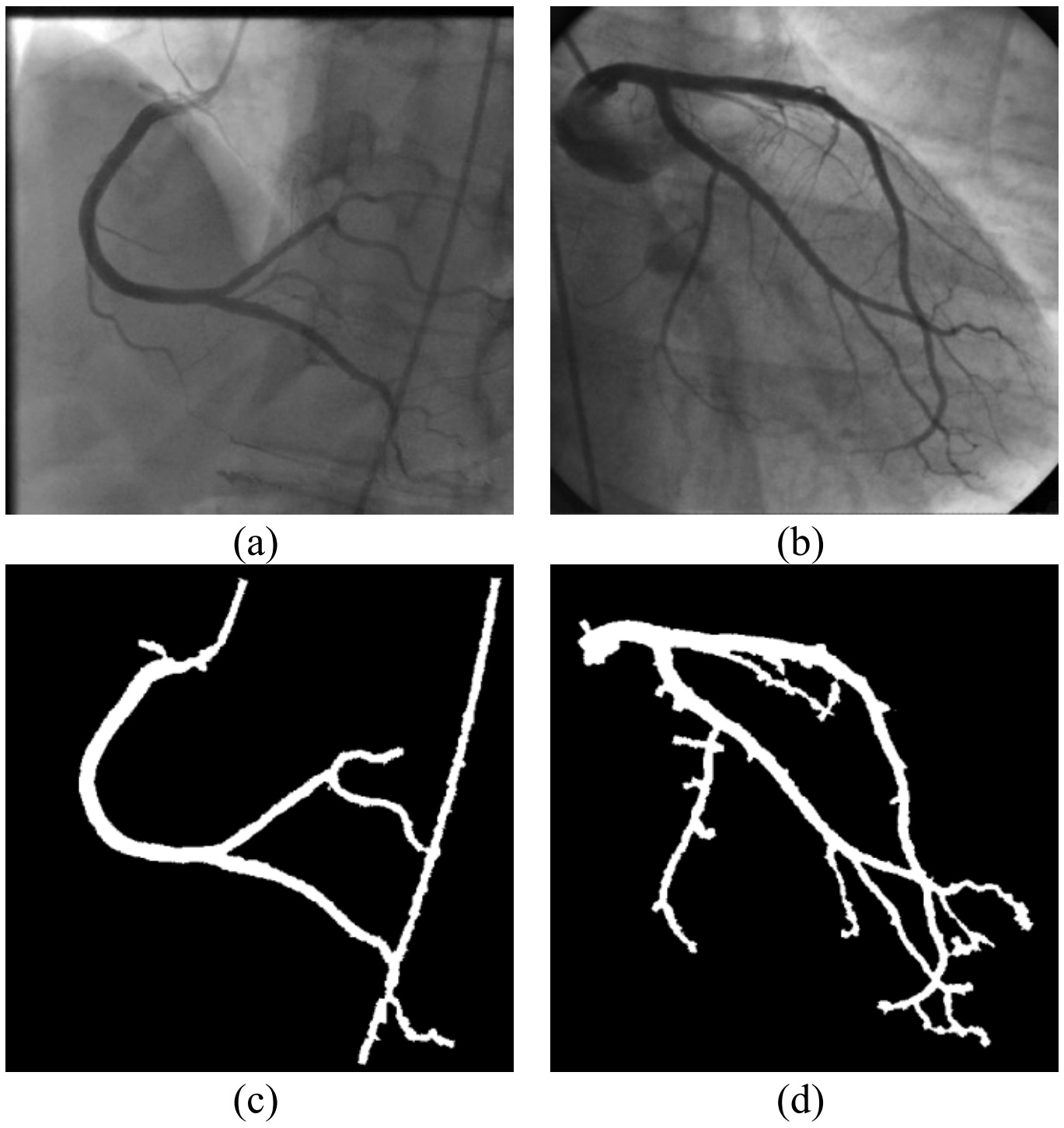

Fig. 10. Results of initial segmentation. (a), (b) Original XRA images. (c), (d) Initial segmentations results.

*2) Segmentation Refinement*

In order to remove the background parts from the initial segmentation result, we process the initial result using orthogonal lines on arteries. For this purpose, the ridge map obtained in the ridge detection stage using a high threshold ($T_H$) is filtered using initial segmentation result. By doing this, the extra ridges are removed.

As illustrated in Fig. 11, for finding the orthogonal lines on arteries, we traverse the ridges and on each ridge pixel $O$ a circle $C$ with the diameter of $d$ is considered that is centered on $O$. For this circle, all the diameters, with different direction from 1 to 180 degree, are determined. Then the average intensities of these diameters are computed. In Fig. 11, the line $R$ is the ridge line and the diameter $L$ is the orthogonal line. As the arteries are darker than the background and the orthogonal diameter has the least number of pixels in the artery, the diameter with the highest average intensity is determined as the orthogonal diameter. In Fig. 12, the result of applying this proposed algorithm for finding orthogonal line is shown.

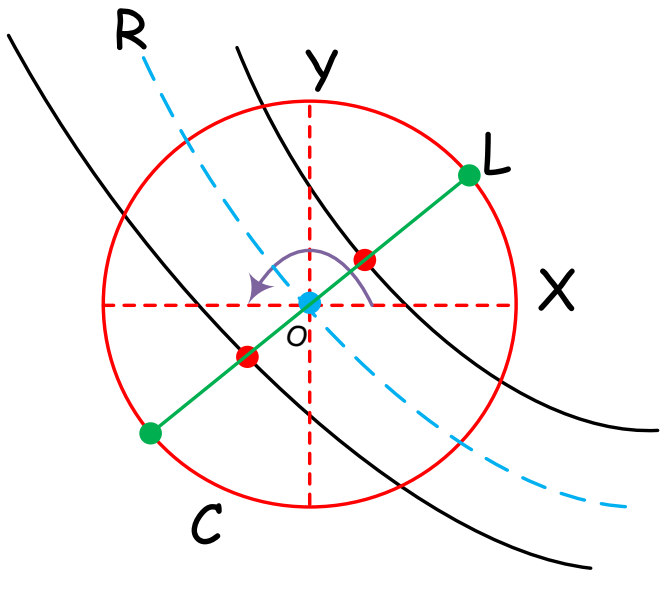

Fig. 11. Results of initial segmentation.

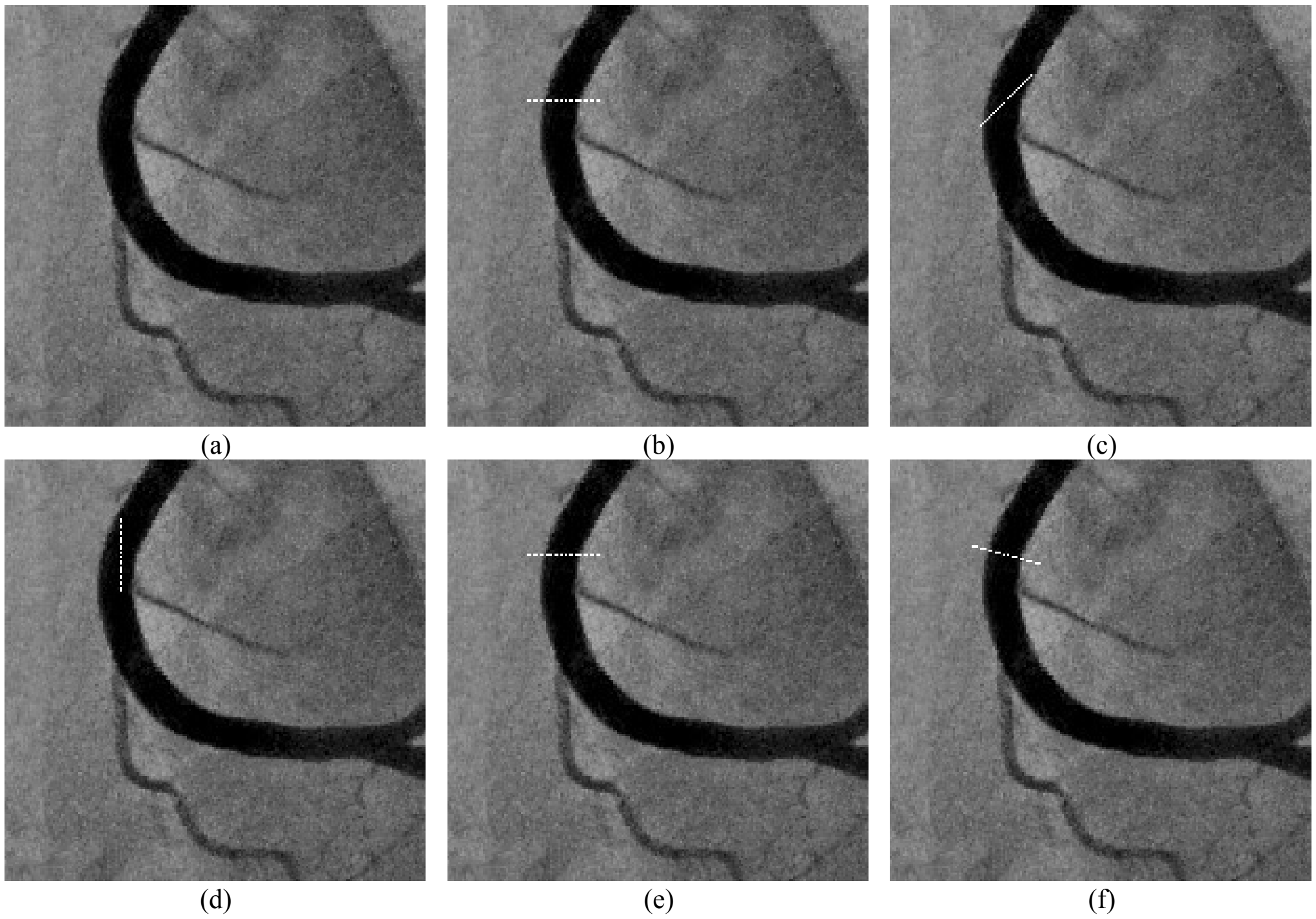

Fig. 12. Result of finding orthogonal line. (a) Original XRA image. (b) 1° Diameter. (c) 45° Diameter. (d) 90° Diameter. (e) 180° Diameter. (f) The orthogonal diameter with $165^{o}$

In Fig. 12 the diameters are the white lines. For example, the diameter with $1^{o}, 45^{o}, 90^{o}$ and $180^{o}$ are shown and the diameter with $165^{o}$ is detected successfully as the orthogonal line.

Because of the presence of noise in the contrast enhanced image, the intensities of orthogonal line pixels are smoothed using an averaging filter. We need to further remove the extra ridge pixels that are still present. Hence, we filter these ridge pixels by considering pixel intensities that exist on the lines orthogonal to these pixels. As the orthogonal lines pass across the arteries at each ridge pixel, these lines should have a inverse Gaussian profile. Such Gaussian profile has a minimum in the middle and two maximums, one in each side as illustrated in Fig. 13. This means that the differences between the minimum and the two maximums should be higher than a value. Otherwise the profile most likely does not belong to an artery. Therefore $L_1$ and $L_2$ shown in Fig. 13 should be higher than a minimum distance threshold $T_d$ As there may be some exceptions in some XRA images, in which the contrast of the arteries are generally low, we consider another criterion that may keep some ridge pixels that are on arteries and removed using $T_d$ threshold. For this aim, at first the intensity of contrast enhanced image pixels corresponding to ridge pixels are sorted in ascending order. Then the average of the fourth quarter of these intensities is computed. The ridge pixels, which their corresponding pixels in the contrast enhanced image are higher than this average value, are kept as artery ridges. This adaptive threshold will be more robust to XRA image quality differences.

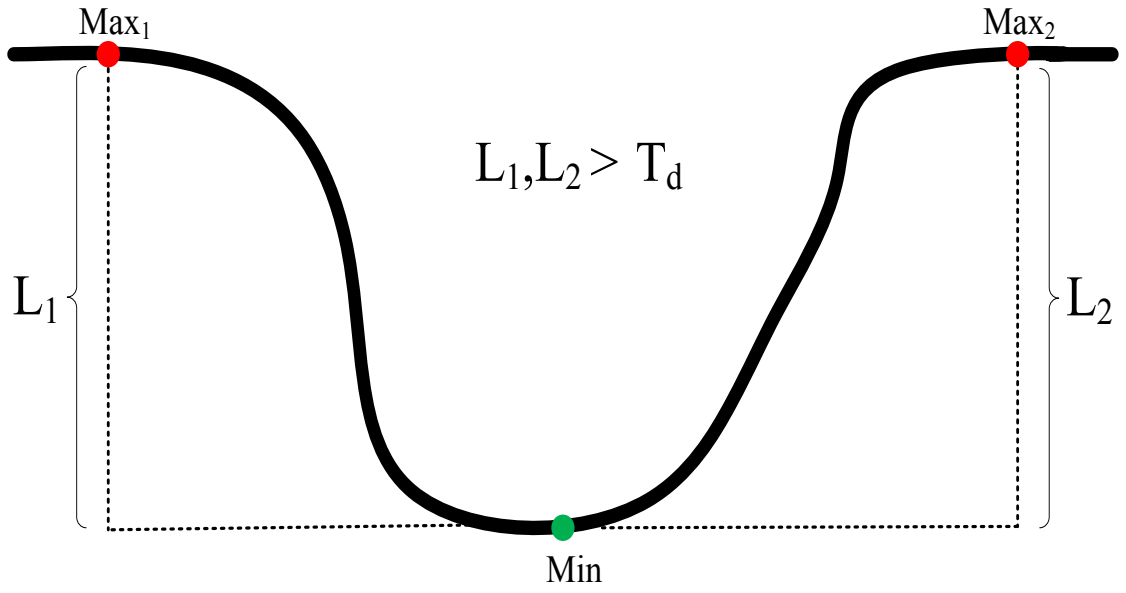

Fig. 13. A sample Gaussian profile.

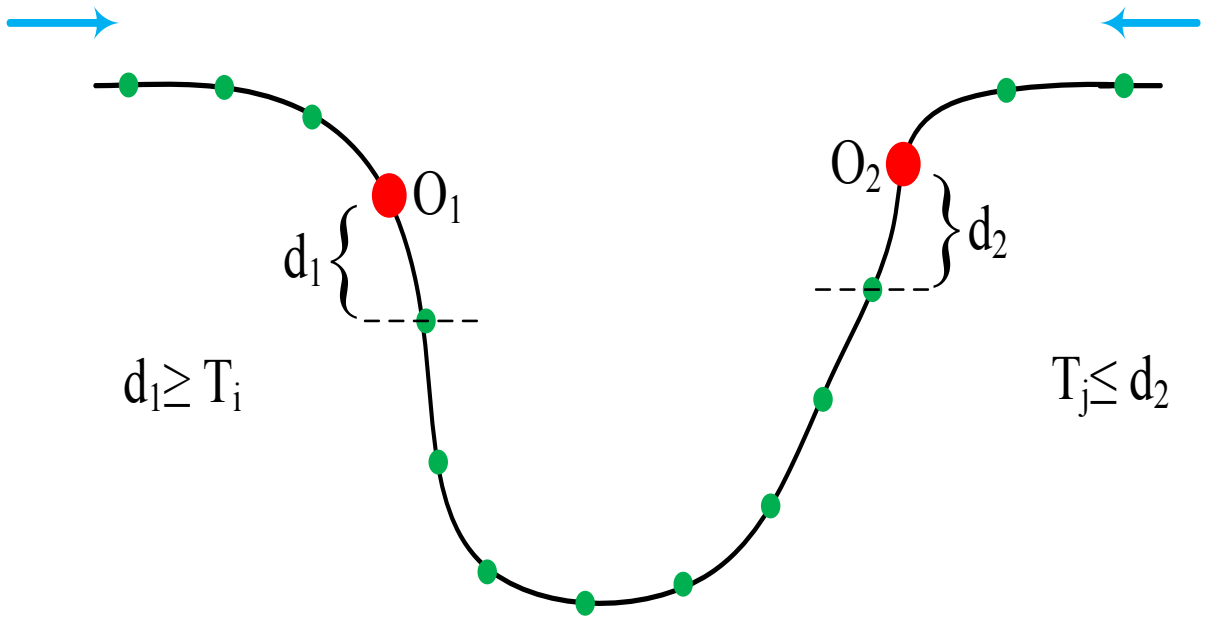

Fig. 14. Finding the arteries boundaries.

After finding the boundary pixels on two sides of a ridge pixel, the minimum of distances between the ridge pixel and the two boundary pixels is computed as $r_d$. Then, a circle centered on the ridge pixel with the radius of $r_d$ is drawn. By doing this a mask for arteries is obtained. By multiplying this mask by the previous mask obtained in the last step using superpixels and ridges (initial segmentation mask), the final arteries. After removing the extra ridge pixels, on the remaining pixels, the Gaussian profiles of the orthogonal lines are processed in order to find the two boundaries of the arteries on each ridge pixel. This process is illustrated in Fig. 14. In this procedure, we begin the search from each side of the profile and compute the difference between the current pixel and the next pixel. If the difference is larger than or equal to a pre-defined threshold, the current pixel is considered as the boundary pixel, otherwise the threshold value is reduced and a new search is performed. As shown in Fig. 14, $T_i$ and $T_j$ are two different thresholds satisfied by $d_1$ and $d_2$ respectively.

mask is formed. In Fig. 15, the result of this refinement is shown. By comparing Fig. 15(c) and Fig. 15(d) with the initial segmentation results in Fig. 10(c) and Fig. 10(d), it can be seen that the proposed refinement procedure has removed the extra background parts that were counted as arteries segment.

*3) Centerline Extraction*

In order to find the arteries centerline, we use the arteries mask obtained in the former step. By multiplying the arteries mask by the low threshold ($T_L$) ridge map, the extra ridges are removed and the ridges corresponding to the arteries are remained which are arteries centerlines. In Fig. 15(e) and Fig. 15(f) the final result of centerline extraction is shown. By considering the segmented arteries, it can be observed that the centerlines of the extracted arteries are extracted properly.

## 5. Results

In this section, we present the evaluation results of our proposed catheter detection and tracking and coronary arteries segmentation methods. In multi-scale Top-Hat transform, the structuring elements were disk shape with varying sizes from 3 to 19. Also, window radius $r$ and regularization parameters $\epsilon$ in the guided filter were set as 8 and 0.2 respectively. The low, medium and high thresholds for ridge detection method were taken as 0.2, 0.25 and 0.4 respectively. The three superpixels scales were taken as 2000, 3000 and 4000. The $T$ threshold for differentiating between vessels and background superpixels was set as 0.5. The $T_d$ threshold in the refinement step was set to 0.2 and the $d$ diameter was set to 25 pixels.

In Fig. 16 and Fig. 17, we present the qualitative and quantitative evaluation results of our proposed catheter detection and tracking method. In Fig. 18 and 19 as well as Tables 1 and 2, we compare our coronary arteries segmentation method qualitatively and quantitatively with one of the state of the art methods [15], which is based on a Graph-cut algorithm. We also present the comparison of the time complexity of our proposed method and [15] in subsection 6.E. For this aim, two challenging datasets, DS1 and DS2 consisting 164 and 298 images respectively, were collected including 8-bit images with 512×512 pixels. In all of these images, different kinds of artifacts such as non-uniform illumination, low contrast, other body organs and low SNR and etc. exist. DS1 includes XRA images in which just the arteries are present or if the catheter is present it has no overlap with the arteries, while DS2 includes XRA images in which in addition to the existence of the arteries, the catheters are also present and have overlaps with the arteries.

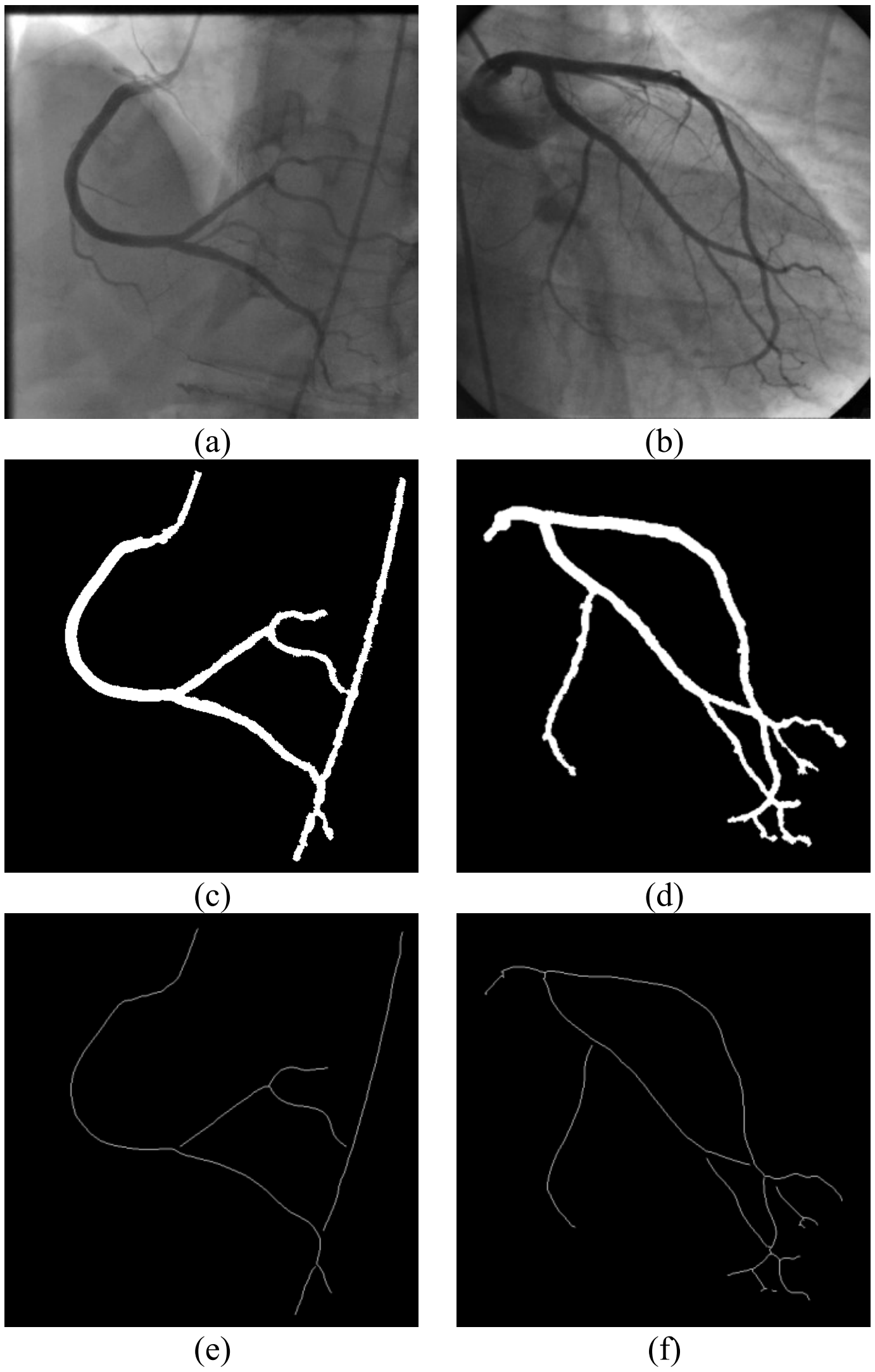

Fig. 15. Final results of segmentation and centerline extraction. (a), (b) Original XRA images. (c), (d) Final segmentations results. (e), (f) Final centerline extraction results.

## 6. Discussions

### *A. Qualitative Evaluation of the Catheter Detection*

The results of applying the proposed catheter detection and tracking method are shown on four different XRA sequences in Fig. 16. In Fig. 16, each sequence is shown in a row. Their first frame is shown in the first column. The other four columns consist of four different phases including before injection, beginning of injection, fully injection, end of injection. In all of these frames the detected catheters are shown in red. It can be seen that the proposed method successfully tracks the catheter. Even the presence of arteries and their overlapping with the catheter do not affect the performance of our method. Also, in these sequences catheter displaces and these displacements have not blundered our method. These qualitative results prove the high effectiveness and accuracy of our method.

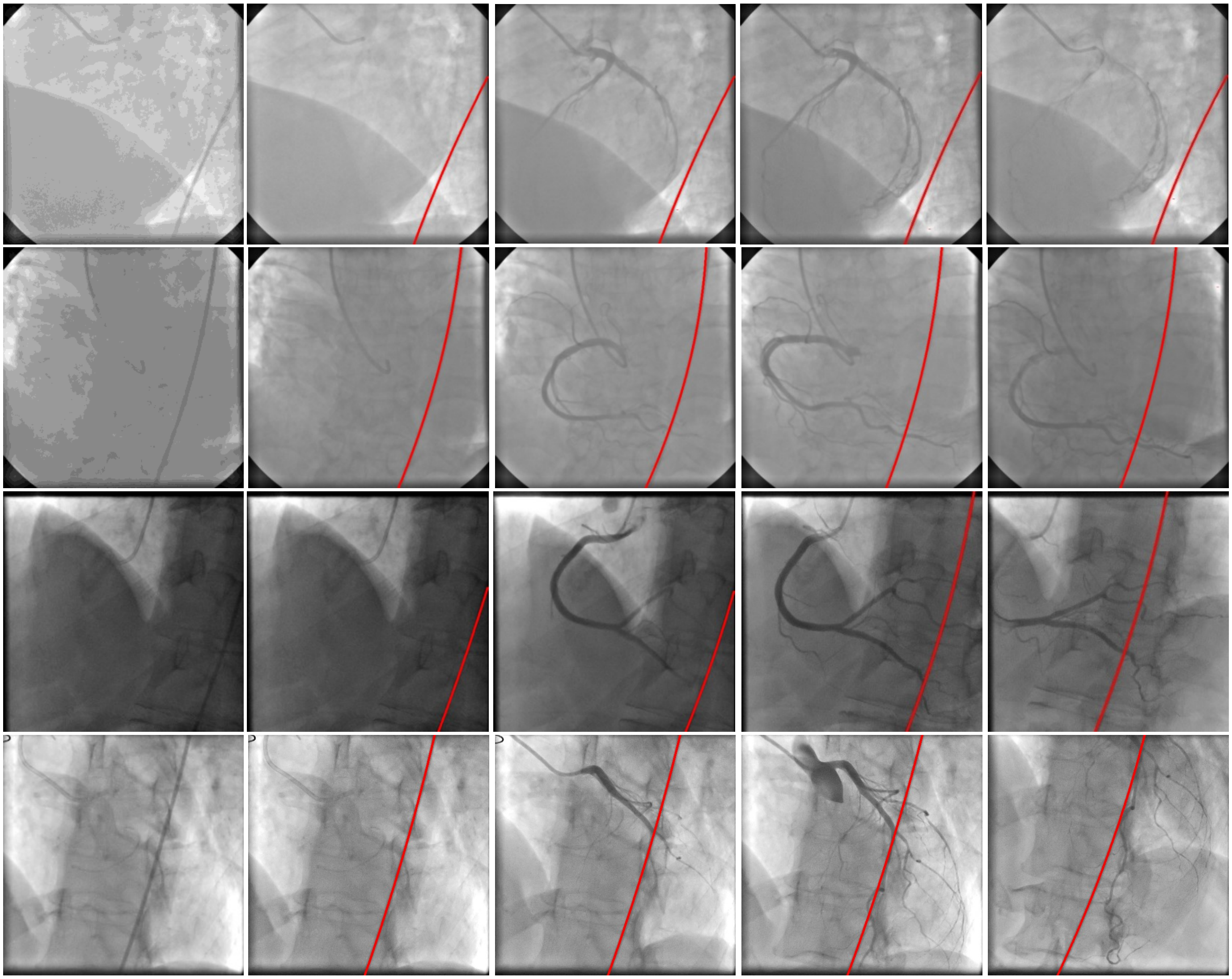

(a) (b) (c) (d) (e)

Fig. 16. The qualitative evaluation of the proposed catheter detection and tracking method. (a) Original first frame, (b) before injection, (c) beginning of injection, (d) full injection, (e) end of injection.

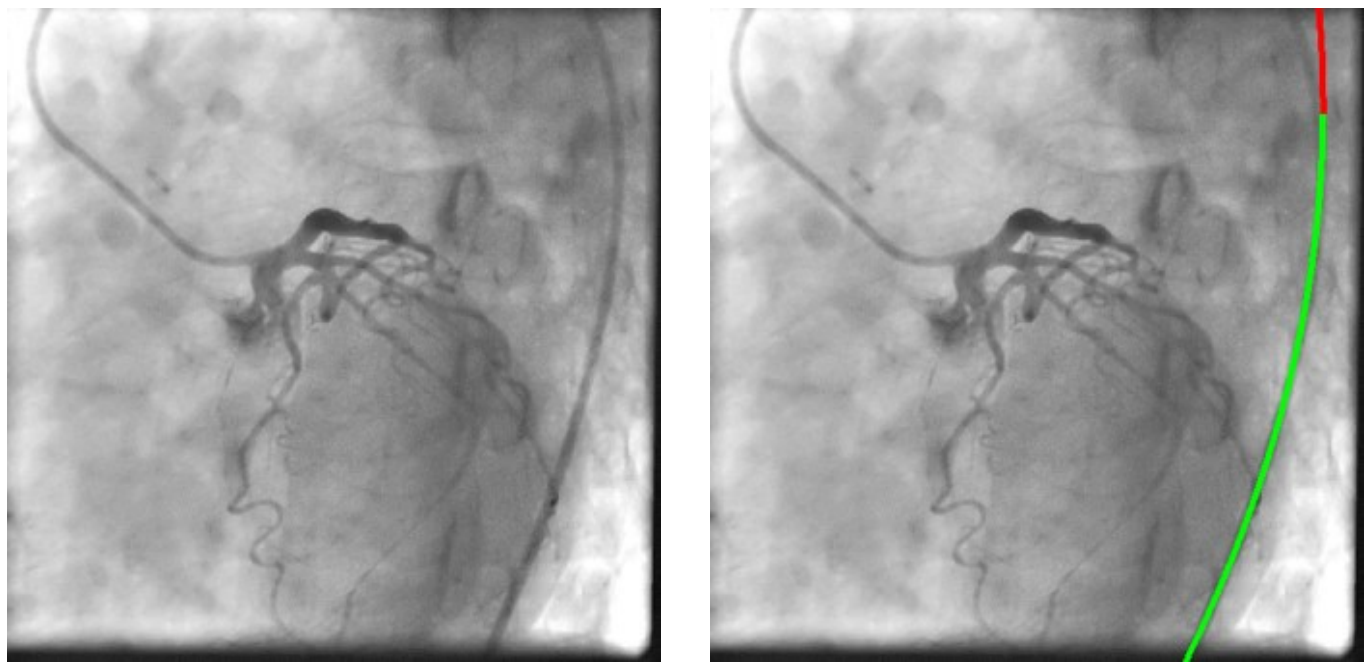

Fig. 17. Catheter tracking precision computing procedure.

### *B. Quantitative Evaluation of the Catheter Detection*

In order to quantitatively evaluate our proposed catheter detection method, we compute the precision on 25 challenging XRA sequences containing catheter. For computing the precision in each frame, we count the pixels of second order polynomial that do not lie on the catheter. An example of this counting procedure is illustrated in Fig. 17. In this figure, green pixels are those that are correctly counted (true positive) and the red pixels are incorrectly considered (false positive). The precision of our method on these sequences turned out to be 0.9597. This high value of precision demonstrates the effectiveness of our method in XRA sequences. Detection and tracking of catheter in such sequences are very challenging.

### *C. Qualitative Evaluation of the Proposed Segmentation Method*

In Fig. 18 and Fig. 19 the qualitative results of the proposed segmentation method and method of [15] are compared. Images in Fig. 18 were selected from DS1. In the first row of Fig. 18 the original XRA images are shown. In the second row the segmentation results of method of [15] are shown in green and in the third row the results of proposed segmentation method are shown in red. The extracted centerlines are shown in green in our results. By comparing the proposed segmentation results with method [15], in Fig. 18, it can be seen that the proposed method has properly segmented the major parts of the arteries but [15], has some faults in detection of arteries. For example, in Fig. 18, in column (d), method [15], cannot find the arteries. Also, in column (e), (f), (g) and (h), method of [15] has not completely segmented the main arteries. On the other hand, the proposed method has completely segmented the major arteries which are vital for stenosis detection. For instance, in column (h) our segmentation method has segmented the arteries even in the non-uniform illumination which has prevented method [15], from segmenting the whole structure of the arteries.

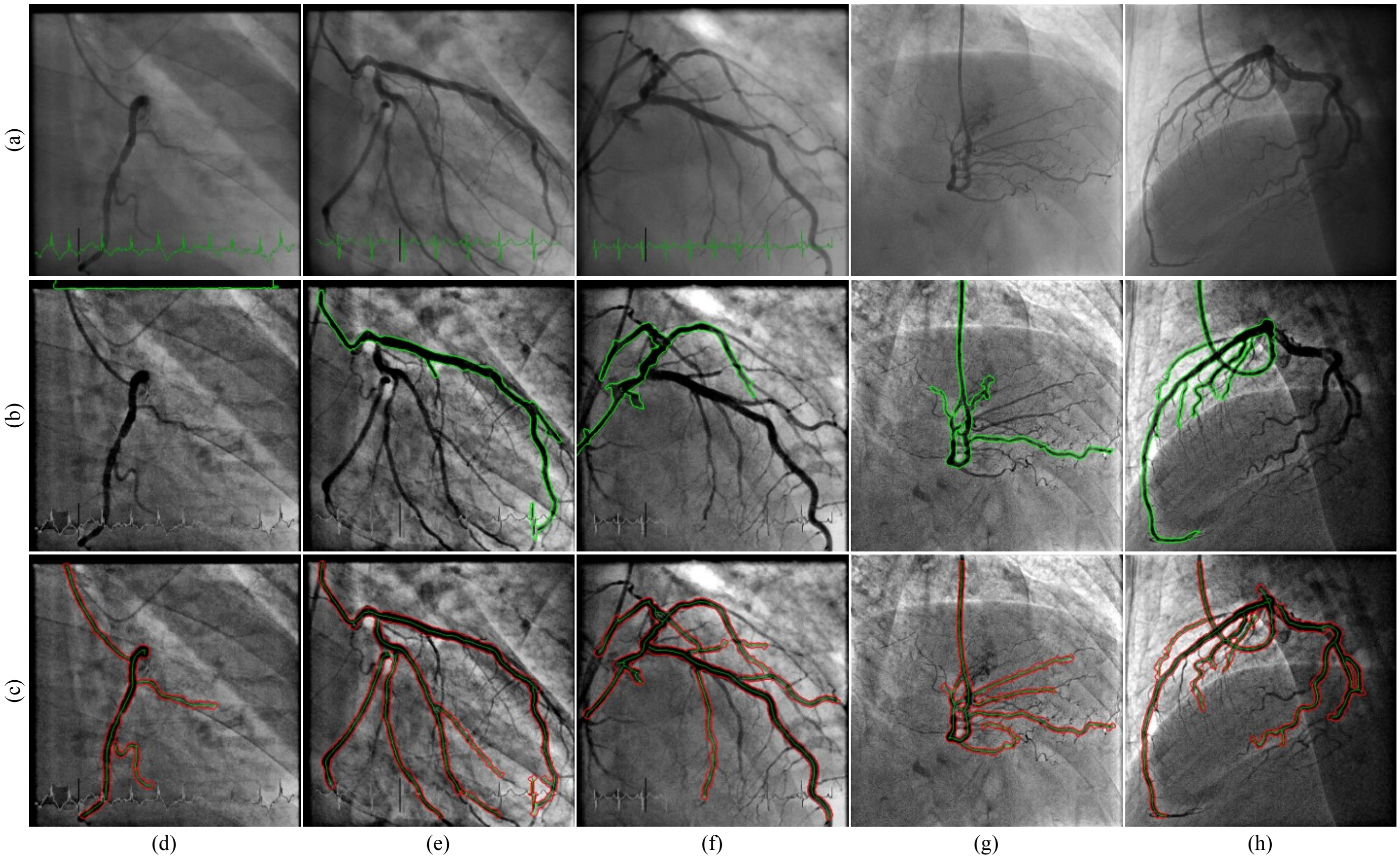

Fig. 18. Qualitative comparison of proposed method and [15], on several DS1 XRA images. (a) Original XRA images. (b) Segmentation results using [15]. (c) Segmentation results using the proposed method.

Also in Fig. 19 the comparisons between our method and [15], on XRA images from DS2 are presented. In the first row of Fig. 18 original images are shown, in the second row the segmentation results of [15], are shown in green and in the third row the results of our proposed arteries segmentation and catheter detection method and the extracted centerlines are shown in red, blue and green respectively. In all of these images, method [15], like other coronary arteries segmentation methods, detects the catheter as arteries and this detection adds a great false positive value to the segmentation results which is indicated in the quantitative evaluations.
Furthermore, in Fig. 19 in all columns from (d) to (h), the graph-cut based method [15], has detected the catheter as the arteries. Also in column (e) and (f) method [15], has just segmented the catheter and was not able to detect the arteries. But our proposed segmentation method consists of a catheter detection method which can reduce the false positive of our segmentation method in cases that the catheters exist in XRA frames. Fig. 19 shows the necessity of a catheter detection method appropriately which improves the segmentation results

### *D. Quantitative Evaluation of the Proposed Segmentation Method*

In order to compare our method with [15] the results of both methods were assessed by a cardiologist on DS1 and DS2 datasets. The cardiologist labeled each processed image by one of the following labels:

Label 0: Insufficient - major vessels are missing (not just branches) or major areas without vessels are inappropriately identified as vessels.

Label 1: Limited - major vessels are identified but the results do not include substantial branches or they inappropriately include substantial areas without vessels.

Label 2: Good - major vessels and substantial branches are included; only minor branches that are unlikely to be clinically significant are missing.

Label 3: Excellent - the entire vasculature from major vessels to major and minor branches are included.

The evaluation results for DS1 and DS2 are presented in the Table 1 and Table 2 respectively. By comparing the results shown in Table 1, we notice that our method outperforms method of [15] as none of the results of our method were labeled as 0. Also our proposed method labeled as 2 or 3 more than [15]. These results demonstrate that our method is more reliable. The results in Table 2 show the effectiveness of a catheter detection method in the performance of an artery segmentation method. As shown in Table 2, 37% of the results of [15] on DS2 dataset were labeled as 0. This is in contrast to our method which only has 1% of such labels. On the other hand, none of the results of method [15] were detected as Excellent as it segmented the catheter as an artery.

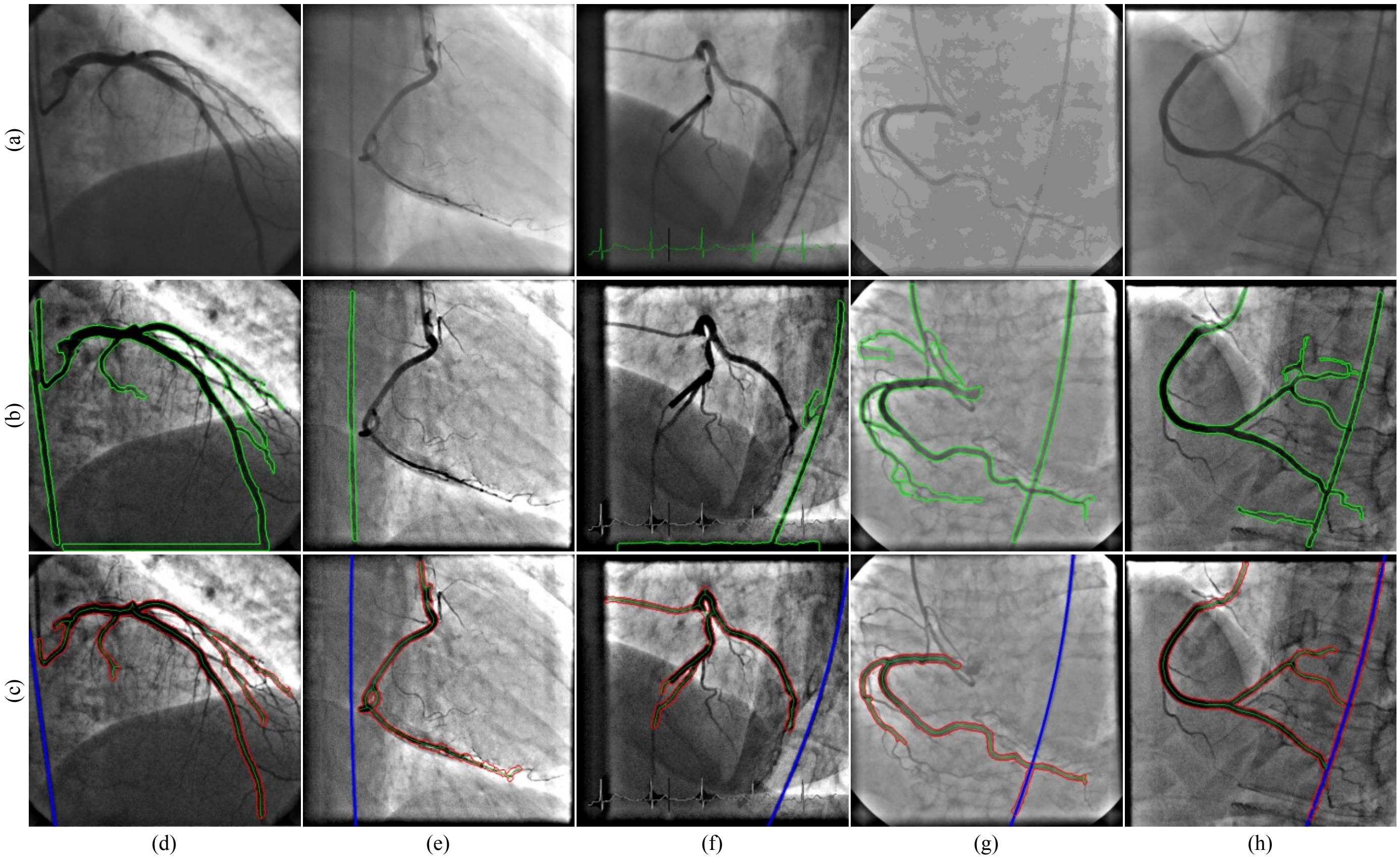

Fig. 19. Qualitative comparison of proposed method and [15], on several DS2 XRA images. (a) Original XRA images. (b) Segmentation results using [15]. (c) Segmentation results using the proposed method.

Table 1. Quantitative Evaluation on DS1 dataset.

| | Insufficient (0) | Limited (1) | Good (2) | Excellent (3) |
|---|---|---|---|---|
| Proposed Method | 0% | 16% | 56% | 28% |
| Method [15] | 24% | 26% | 46% | 4% |

Table 2. Quantitative Evaluation on DS2 dataset.

| | Insufficient (0) | Limited (1) | Good (2) | Excellent (3) |
|---|---|---|---|---|
| Proposed Method | 1% | 15% | 58% | 26% |
| Method [15] | 37% | 16% | 47% | 0% |

### *E. Time Complexity Comparison*

In order to compare the time complexity of the proposed method with [15], we executed both methods in MATLAB on the same system with Core i7 2.6 GHz and 8GB RAM. The proposed method required 67% less time than [15]. It should be noted that some parts of the graph-cut source code in [15], is written in C language which is much faster than a correspondent MATLAB code.

## 7. Conclusion

In this paper a new automatic framework for coronary artery segmentation, catheter detection and centerline extraction was proposed. In this method in the preprocessing stage the input XRA image was processed for better artery segmentation using superpixels and catheter detection and tracking in the following stages. Also for smoothing the input XRA image we exploited the contrast enhanced image and the vesselness map for better smoothing of the background parts. This was done for better preservation of the image contents such as the catheter and arteries.

Also we exploited our previous automatic catheter detection method using image ridges. In this method the catheter ridge was detected in the first frame of a sequence and a second polynomial was fit on the detected catheter ridge. Then the parameters of the second order polynomial were used in the second frame for detecting the catheter ridge. This procedure was continued throughout the sequence.

The proposed artery segmentation stage was based on superpixel processing. Because of the intensity differences of artery parts and background parts of XRA images, superpixels lay appropriately on the boundaries of arteries. This made the segmentation procedure much faster and easier. In the proposed segmentation method, the SLIC superpixel [15] was applied on the enhanced image and a vesselness probability was obtained for each superpixel. A threshold was determined for detecting the arteries superpixels. Also the superpixels that lay on the ridges that were common with arteries regions were added.

We also extracted the arteries centerlines using the arteries mask. By multiplying this mask with the image ridges, arteries centerlines were extracted. By obtaining the catheter mask, the arteries mask and centerline mask and overlaying them on the coordinated enhanced image, the final segmented image was produced.

We evaluated our catheter detection method and arteries segmentation method separately qualitatively and quantitatively. The qualitative and quantitative evaluation results of the proposed catheter detection method proved that our method has high precision. Even when arteries were overlapping with the catheter or when body and camera motions were present our method could track the catheter. Also we compared our segmentation results with a state-of-the-art method [15]. The result showed that our segmentation results in the frames that catheter was not present, could segment the main arteries better than [15]. Also, in the frames that catheter existed or it had overlaps with the arteries structure, our method performed better. Our method could detect the catheter and reduced the false positive rate in addition to achieving better segmentation results. Method of [15], similar to other coronary arteries segmentation methods, segmented the catheter as an artery or in some cases just segments the catheter instead of arteries.

## Acknowledgment

The authors would like to thank Sina Heart Center, Isfahan, Iran for providing us with angiogram videos and also Dr. Antonio Hernández Vela from Universitat de Barcelona for sharing the source code of vessel segmentation [15], with us.

## References

[1] S. Kato *et al.* (2010). Assessment of coronary artery disease using magnetic resonance coronary angiography: a national multicenter trial. *Journal of the American College of Cardiology, 56*(12), 983-991.

[2] H. R. Fazlali, N. Karimi, S. M. R. Soroushmehr, S. Samavi, B. Nallamothu, H. Derksen, and K. Najarian (2015). Robust catheter identification and tracking in X-ray angiographic sequences. In *Engineering in Medicine and Biology Conference* (pp. 7901-7904).

[3] A. F. Frangi, W. J. Niessen, K. L. Vincken, and M. A. Viergever (1998). Multiscale vessel enhancement filtering. In *Medical Image Computing and Computer-Assisted Intervention* (pp. 130-137).

[4] P. T. Truc, M. A. Khan, Y. K. Lee, S. Lee, and T. S. Kim (2009). Vessel enhancement filter using directional filter bank. *Computer Vision and Image Understanding, 113,* 101-112.

[5] M. A. Khan, M. K. Khan, and M. A. Khan (2004). Coronary angiogram image enhancement using decimation-free directional filter banks. In *IEEE International Conference on Acoustics, Speech, and Signal Processing* (pp. 441-444).

[6] H. R. Fazlali, N. Karimi, S. M. R. Soroushmehr, S. Sinha, S., Samavi, B. Nallamothu, and K. Najarian (2015). Vessel region detection in coronary X-ray angiograms. In *International Conference on Image Processing* (pp. 1493-1497).

[7] S. Petkov, X. Carrillo, P. Radeva, and C. Gatta (2014). Diaphragm border detection in coronary X-ray angiographies: New method and applications. *Computerized Medical Imaging and Graphics, 38*, 296-305.

[8] H. Ma, A. Hoogendoorn, E. Regar, W. J. Niessen, T. van Walsum, Automatic online layer separation for vessel enhancement in X-ray angiograms for percutaneous coronary interventions. *Medical Image Analysis,* 39, 145-161, 2017.

[9] S. Tang, Y. Wang, and Y. W. Chen (2012). Application of ICA to X-ray coronary digital subtraction angiography. *Neurocomputing, 79*, 168-172.

[10] M. Nejati, and H. Pourghassem (2014). Multiresolution image registration in digital X-ray angiography with intensity variation modeling. *Journal of medical systems*, *38*(2), 1-10.

[11] C. Kirbas, and F. K. Quek (2003). Vessel extraction techniques and algorithms: a survey. In *IEEE Symposium on Bioinformatics and Bioengineering* (pp. 238-245).

[12] B. Felfelian, H. R. Fazlali, N. Karimi, S. M. R. Soroushmehr, S. Samavi, B. Nallamothu, K. Najarian. Vessel segmentation in low contrast X-ray angiogram images. International Conference on Image Processing (ICIP), pp. 375-379, 2017.

[13] F. M'hiri, N. L. T. Hoang, L. Duong, and M. Cheriet (2012). A new adaptive framework for tubular structures segmentation in X-ray angiography, In *International Conference on Information Science, Signal Processing and their Applications* (pp. 496-500).

[14] M. Taghizadeh Dehkordi, A. M. Doosthoseini, S. Sadri, H. Soltanianzadeh (2014). Local feature fitting active contour for segmenting vessels in angiograms. *IET Computer Vision, 8,* 161-170.

[15] A. Hernández-Vela, C. Gatta, S. Escalera, L. Igual, V. Martin-Yuste, M. Sabaté, and P. Radeva (2012). Accurate coronary centerline extraction, caliber estimation, and catheter detection in angiographies. *IEEE Transactions on Information Technology in Biomedicine*, 16, 1332-1340.

[16] R. Achanta, A. Shaji, K. Smith, A. Lucchi, P. Fua, and S. Susstrunk (2012). SLIC superpixels compared to state-of-the-art superpixel methods. *IEEE Transactions on Pattern Analysis and Machine Intelligence, 34*, 2274-2282.

[17] Y. C. Tsai, G. J. Lee, and M. Y. C. Chen (2015). Automatic segmentation of vessels from angiogram sequences using adaptive feature transformation. *Computers in Biology and Medicine, 62*, 239-253.

[18] X. Bai, F. Zhou, and B. Xue (2012). Image enhancement using multi scale image features extracted by top-hat transform. *Optics & Laser Technology, 44*, 328-336.

[19] K. He, J. Sun, and X. Tang, Guided image filtering (2013). *IEEE Transactions on Pattern Analysis and Machine Intelligence, 35*, 1397-1409.

[20] A. M. López, F. Lumbreras, J. Serrat,and J. J. Villanueva (1999). Evaluation of methods for ridge and valley detection. *IEEE Transactions on Pattern Analysis & Machine Intelligence, 21*(4)*,* 327-335.